%% file: main.tex
\documentclass[10pt,twocolumn,letterpaper]{article}

\usepackage[pagenumbers]{cvpr}      

\usepackage{wrapfig}
\usepackage[dvipsnames]{xcolor}
\usepackage{multirow,mathtools } 
\usepackage{algorithm,algpseudocode}
\usepackage{pifont}
\usepackage{color, colortbl}
\usepackage{blindtext}
\usepackage{lipsum}
\usepackage{multirow}
\usepackage{graphicx}
\usepackage{listings}
\usepackage[most]{tcolorbox}
\usepackage{booktabs} 
\usepackage[table]{xcolor}
\usepackage{caption}
\usepackage{amsmath}
\usepackage{amssymb}

\definecolor{cvprblue}{rgb}{0.21,0.49,0.74}
\usepackage[pagebackref,breaklinks,colorlinks,allcolors=cvprblue]{hyperref}

\def\paperID{19733} 
\def\confName{CVPR}
\def\confYear{2026}

\title{Flash-DMD: Towards High-Fidelity Few-Step Image Generation with Efficient Distillation and Joint Reinforcement Learning}

\def\spaces{~~~~~}
\author{Guanjie Chen\textsuperscript{1,2}\footnotemark[1]\spaces{}
Shirui Huang\textsuperscript{2}\footnotemark[1]\spaces{}
Kai Liu\textsuperscript{2}\spaces{}
Jianchen Zhu\textsuperscript{2}\spaces{} \\
Xiaoye Qu\textsuperscript{3}\spaces{}
Peng Chen\textsuperscript{2}\spaces{}
Yu Cheng\textsuperscript{4}\footnotemark[2]\spaces{}
Yifu Sun\textsuperscript{2}\footnotemark[2]\\\\
\textsuperscript{1}Shanghai Jiao Tong University ~~
\textsuperscript{2}Tencent \\
\textsuperscript{3}Huazhong University of Science and Technology ~~
\textsuperscript{4}The Chinese University of Hong Kong \\
\textit{\small chenguanjie@sjtu.edu.cn}
~ \textit{\small yifusun@tencent.com}
}

\newcommand{\ours}{Flash-DMD}

\begin{document}

\twocolumn[{%
\renewcommand\twocolumn[1][]{#1}%
\maketitle
\begin{center}
    \centering
    \captionsetup{type=figure}
    \vspace{-20pt}
    \includegraphics[width=0.95\linewidth]{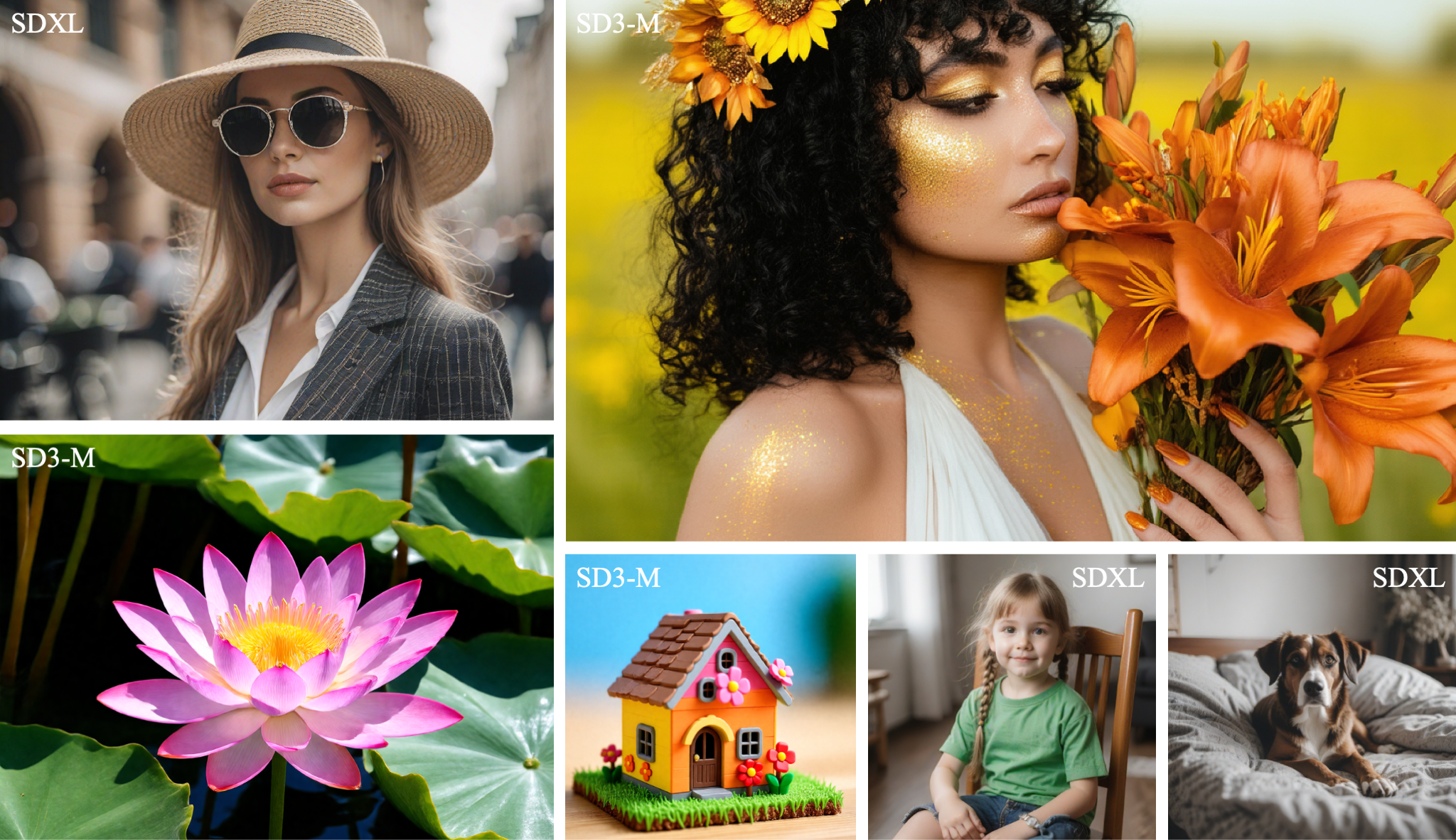}
    \vspace{-8pt}
   \caption{Samples from 4-step \ours~ on SDXL and SD3-Medium. \ours~ takes less than $3\%$ training cost of DMD2~\cite{DMD2}.}
    \label{fig:teaser}
\end{center}%
}]

\renewcommand{\thefootnote}{\fnsymbol{footnote}}
\footnotetext[1]{Equal Contribution.} 
\footnotetext[2]{Corresponding Authors.}
\input{sec/0_abstract}  
\input{sec/1_intro}

\input{sec/2_relate}
\input{sec/3_method}
\input{sec/4_exp}
\input{sec/5_conclude}

{
    \small
    \bibliographystyle{ieeenat_fullname}
    \bibliography{main}
}

\input{sec/X_suppl}

\end{document}

%% file: sec/0_abstract.tex
\begin{abstract}
Diffusion Models have emerged as a leading class of generative models, yet their iterative sampling process remains computationally expensive. 
Timestep distillation is a promising technique to accelerate generation, but it often requires extensive training and leads to image quality degradation. 
Furthermore, fine-tuning these distilled models for specific objectives, such as aesthetic appeal or user preference, using Reinforcement Learning (RL) is notoriously unstable and easily falls into reward hacking. 
In this work, we introduce Flash-DMD, a novel framework that enables fast convergence with distillation and joint RL-based refinement.
Specifically, we first propose an efficient timestep-aware distillation strategy that significantly reduces training cost with enhanced realism, outperforming DMD2 with only $2.1\%$ its training cost. Second, we introduce a joint training scheme where the model is fine-tuned with an RL objective while the timestep distillation training continues simultaneously. We demonstrate that the stable, well-defined loss from the ongoing distillation acts as a powerful regularizer, effectively stabilizing the RL training process and preventing policy collapse. 
Extensive experiments on score-based and flow matching models show that our proposed Flash-DMD not only converges significantly faster but also achieves state-of-the-art generation quality in the few-step sampling regime, outperforming existing methods in visual quality, human preference, and text-image alignment metrics. Our work presents an effective paradigm for training efficient, high-fidelity, and stable generative models. Codes are coming soon.
\end{abstract}

%% file: sec/1_intro.tex
\input{figures/main_figure}
\section{Introduction}

Diffusion models \citep{diffusion, latentdiffusion, sdxl, flowdiffusion, flux2024} have demonstrated remarkable success in text-to-image generation in recent years. 
Numerous iterative denoising steps poses a significant obstacle to real-time or resource-constrained deployment. 
To address this issue, various diffusion distillation techniques have been developed to distill multi-step teacher diffusion models into efficient student models that can produce comparable image quality in just one or a few inference steps \citep{LCM, PCM, DMD, DMD2, sdxllighting, flashsdxl, senseflow, ADM}. 
However, most existing distillation methods require thousands of GPU hours for training.
This significantly limits its accessibility to research groups and institutions with limited resources, and hinders rapid deployment in practical applications. 

\noindent
Among existing distillation methods, Distribution Matching Distillation (DMD) methods \citep{DMD, DMD2, ADM, senseflow} stand out for their superior generative quality, leveraging variational score distillation objectives \citep{scoredistill} to align the output distributions of student and teacher models. However, this objective function suffers from unstable training and a tendency to mode seeking. 
Some approaches have employed adversarial methods to mitigate these problems. DMD2 \citep{DMD2} proposes latent adversarial regulations with real images and designs a Two-Time scale Update Rule (TTUR) to stabilize training, but it combines the GAN \citep{ADD, LADD} framework with DMD in a naive manner, neglecting the timestep aware feature of timestep distilled diffusion models, and the fake score $\mu_{fake}$ is trained both to discriminate real and generated images, but also to track the distribution changes of student models. These design compromises its efficiency in matching the distribution of teacher models. Furthermore, DM loss is inefficient in the latter part of distillation as it is hard to guide detailed learning, preventing it from effectively guiding the student diffusion model. These observations motivate our core research questions:

\begin{tcolorbox}[before skip=0.2cm, after skip=0.2cm, boxsep=0.0cm, middle=0.1cm, top=0.1cm, bottom=0.1cm]
\textit{\textbf{Q1} In the early phase, how can we more effectively coordinate distribution matching with perceptual realism enhancement to accelerate convergence? \\
\textbf{Q2} In the later phase, how can we more effectively refine the student model for better visual details and perceptual fidelity in a direct way?}
\end{tcolorbox}

\noindent
To address the inefficiencies of the distribution matching methods, we proposed \ours, a twofold method in the few-step distillation task that outperforms DMD2 with a much smaller training cost, while achieving superior perceptual realism. Specifically, our \ours~follows different principles in the early and later generation phases. 
\noindent
\textit{In the early phase}, as the denoising performance at different timesteps varies, the distillation target should also differ. 
To this end, we decouple the adversarial training and distribution matching frameworks with timestep aware strategy. 
Specifically, at high-noise timesteps, the denoising model's primary objective is to learn global composition and structure from the teacher. Considering DM loss is effective to process noisy latents, we use a pure DM loss to align the student model with the teacher model's output distribution. 
At low-noise timesteps, the model focuses on refining fine-grained details and enhancing perceptual realism. Thus, we use Pixel-GAN to match the distribution of real images and improve the photorealism of the generated images. 
\textit{In the later phase}, we further optimize generation quality to align with human preferences. Previous reinforcement learning works on few-step distilled models \cite{pso, hypersd} suffer from the reward hacking phenomenon, which produces ``oil painting" artifacts. 
We combine the Distillation framework with latent reinforcement learning designed especially for few-step models to refine the their handling of fine-grained details efficiently. The framework can effectively alleviating reward hacking problem.

\noindent
By combining faster convergence in the early phase with joint finer optimization in the latter phase, we demonstrate the efficiency and superior performance of our method of distilling from SDXL to produce high-quality, realistic images. In particular, our method achieves the highest human preference scores while requiring the lowest training cost in DMD series methods to date.

\noindent
To summarize, our main contributions are threefold:

\begin{itemize}

\item  At the first stage, we decouple the training objectives of Distribution Matching series via a timestep-aware
strategy to efficiently distill the fundamental distribution of the teacher model in low-SNR timesteps
and refine perceptual quality and texture in high-SNR timesteps, and we counteract the mode-seeking of the DM loss with Pixel-GAN that robustly enhances realism. We also improve the score estimator for fastest convergence and stabilized distillation. In this stage, we achieve the best performance with only $2.1\%$ training cost of DMD2.

\item At the next stage, we design reinforcement learning specifically for the distilled model and integrate it into the distillation process. These innovations eliminate the need for separate reinforcement and distillation phases, significantly reducing computational training costs, avoiding reward-hacking and achieving the best fine-grained details and perceptual fidelity in few steps of generation.

\item By combining stages 1 and 2, we propose \ours. Extensive experiments demonstrate that our method achieves superior performance compared to both the teacher model and baseline in terms of image quality, human preference, and text-image alignment metrics. Our method exhibits strong generalization ability on both score-based diffusion and flow matching models.
\end{itemize}

%% file: figures/main_figure.tex
\begin{figure*}[t]
\centering
\includegraphics[width=\linewidth]{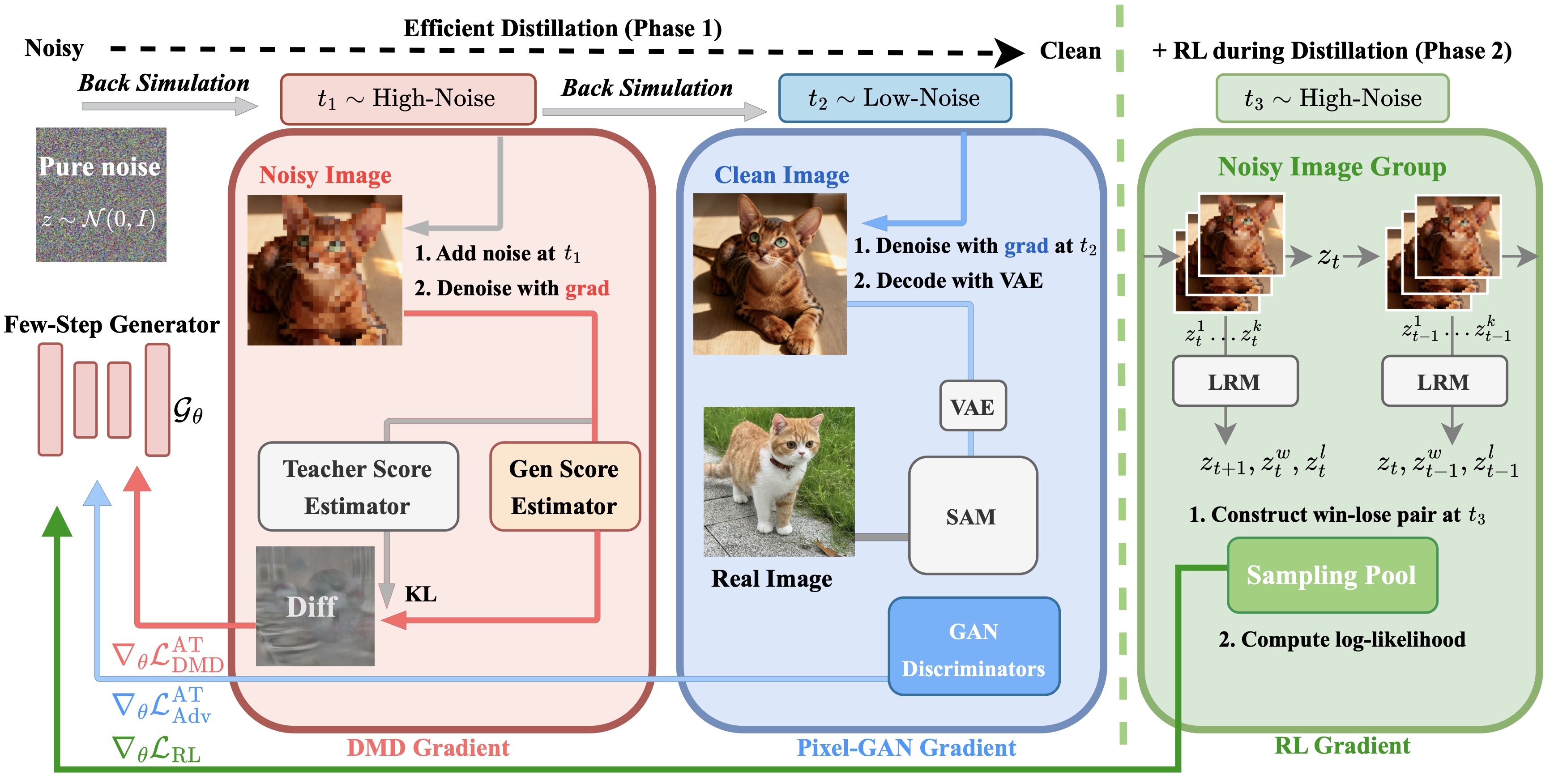}
\caption{
Overview of our proposed \ours. 
We decouple the distillation objective by timestep into a Diffusion Matching loss and an adversarial loss. During high-noise timesteps, the DMD loss enables rapid alignment with the teacher model, while at low-noise timesteps and on real images, Pixel-GAN loss is employed to enhance realism and texture details. This design achieves a more efficient distillation. Building upon this, we further introduce a reinforcement strategy specifically tailored for few-step distilled models, which seamlessly integrates with the distillation objective to achieve superior and more stable performance.
}
\label{fig:my_pdf_figure}
\end{figure*}

%% file: sec/2_relate.tex
\section{Related Work}

\paragraph{Diffusion Models.}
Diffusion models\citep{diffusion, latentdiffusion, song2019generative, dit, skipdit} are a powerful family of generative models that have demonstrated state-of-the-art performance across diverse generative tasks. In text-to-image task, diffusion models operate through two primary stages: a forward process and a reverse process. The forward process disturbs the real image $\mathbf{x}_0 \sim \mathcal{P}_{real}$ with noise $\epsilon \sim \mathcal{N}(0, \mathbf{I})$ by a stochastic equation, at each timestep $t \in \{1\ldots T\}$:
\begin{equation}
\mathbf{x}_t = \sqrt{\alpha_t} \mathbf{x}_{t-1} + \sqrt{1 - \alpha_t} \mathbf{\epsilon}_{t-1}
\end{equation}
where $\alpha$ determines the noise level and finally $\mathbf{x}_T$ will reach $\mathcal{N}(0, \mathbf{I})$. The reverse process then solve the probability flow(SDE) ordinary differential equation to reconstruct $\mathbf{x}_0$. The SDE forward of the denoising diffusion probabilistic model (DDPM) is solved as follows, where $\mu_\theta$ and $\Sigma_\theta$ refer to the predicted mean and covariance:
\begin{equation}
p_\theta(\mathbf{x}_{t-1}|\mathbf{x}_t) = \mathcal{N}(\mathbf{x}_{t-1}; \mu_\theta(\mathbf{x}_t, t), \Sigma_\theta(\mathbf{x}_t, t)),
\end{equation}
The prediction is performed by blocks of transformer\citep{khan2022transformers} or UNet\citep{unet} networks; Generating a complete image requires iterating this reverse step numerous times, making it a time-consuming process. In this work, we focus on distilling the solution of the reverse process more efficiently.

\paragraph{Diffusion Distillation.}

\textit{Progressive Distillation} \citep{salimans2022progressive, progressive, sdxllighting} reduce inference steps in diffusion models by iteratively halving them, ultimately producing a one-step generator. Although effective, this iterative process is computationally expensive and is constrained by the preceding teacher model’s quality, leading to compounding errors.
\textit{Consistency Distillation} \citep{LCM, lcmlora} enforces a consistency constraint for the diffusion models, stipulating that any point on a given trajectory will revert to its starting point. However, it leads to performance degradation in the few-step inference. To migrate the issue, recent works \citep{PCM, hypersd, tcm} have segmented the trajectory and progressively perform distillation on timestep segments.
\textit{Adversarial Distillation} introduces a discriminator to align the few-step student's output with the multi-step teacher's, either at the pixel level \citep{ADD} or latent level \citep{LADD}, DMD2~\citep{DMD2}. DMD2~\citep{DMD2} also uses latent adversarial training to match real-world data distribution, but its straightforward combination of adversarial loss and distribution matching may introduce conflicting objectives that can hinder overall distillation efficiency. 
\textit{Score Distillation} was adapted for the distillation of diffusion models themselves \citep{DMD, nguyen2024swiftbrush, franceschi2023unifying}. An early approach, Distribution Matching Distillation (DMD) \citep{DMD}, aims to minimize the KL-divergence between the teacher and student distributions. DMD2 \citep{DMD2} replaced regression loss with adversarial loss for better realism. Building on this, Adversarial Distribution Matching (ADM) \citep{ADM} introduced a GAN framework with Hinge loss, while SenseFlow \citep{senseflow} optimized scorers and discriminators for efficient distillation of larger models. 
\paragraph{Reinforcement Learning in T2I Generation.}
Reinforcement learning is rapidly migrating to image generation tasks to align large-scale diffusion models with human feedback. Direct Preference Optimization (DPO) \citep{diffusiondpo, spo, capo, inversiondpo, pso, lpo} and Group Relative Policy Optimization (GRPO) \citep{flowgrpo, dancegrpo, mixgrpo, prefgrpo, tempflowgrpo, zhang2025survey} are two popular paradigms. The former methods construct offline or online win-lose pairs and back-propagate the preference order by the Bradley-Terry formed objectives. The latter methods sample a group of images on the SDE/mixed ODE-SDE trajectory, calculate the normalized advantage within the group, and constrain the policy generation direction.
However, current research on performing RL on few-step models remains quite limited. Pairwise Sample Optimization (PSO)\cite{pso} strengthens the relative likelihood margin between the training and reference sets.

%% file: sec/3_method.tex
\section{Methodology}
\subsection{Preliminary}
Given pretrained diffusion model $\mathcal{T}_\phi(x_t, t)$ as teacher model, where $x_t$ is noisy sample at timestep $t\sim\mathcal{U}(\mathbf{1},\mathbf{T})$, DMD~\citep{DMD} and DMD2~\citep{DMD2} distill it into few-step efficient generator $\mathcal{G}_\theta(x_t, t)$ by minimizing the reverse KL divergence between the teacher model's distribution $p_{\tau}$ and the few-step generator's distribution $p_{\text{gen}}$. DMD series methods estimate $p_{\tau}$ through score estimator $\mu_{\tau}(x_t,t)$, and $p_{\text{gen}}$ is tracked with estimator $\mu_{\text{gen}}(x_t,t)$. Score function of the diffused distribution is:
\begin{equation}
        s(x_t, t) =  -\frac{x_t - \alpha_t \mu(x_t, t)}{\sigma_t^2};
\end{equation}
where $\alpha_t, \sigma_t > 0$ are scalars determined by the noise schedule. $s_\text{gen}$ and $s_\tau$ are vector fields that point towards a higher density of distribution. Gradient of Distribution Matching objective w.r.t. $\theta$ is,
\begin{equation}
\nabla_\theta \mathcal{L}_{\text{DMD}} = -\mathbb{E}_{z, t} \left[ s_{\tau}(\mathcal{G}_\theta(\cdot)) - s_{\text{gen}}(\mathcal{G}_\theta(\cdot)) \frac{d\mathcal{G}_\theta(\cdot)}{d\theta} \right],
\label{eq1}
\end{equation}
where $z \sim \mathcal{N}(\mathbf{0}, \mathbf{I})$, $t \sim \mathcal{U}(\mathbf{0}, \mathbf{T})$. In addition to the Distribution Matching objective, DMD2 introduces the combination with adversarial training with real images. Gradient of generator's adversarial objective w.r.t. $\theta$ is,
\begin{equation}
\nabla_\theta \mathcal{L}_{\text{AdvGen}} = \mathbb{E}_{z, t} \left[\log\mathcal{D}\left(\mathcal{G}_\theta(\cdot)\right) \frac{d\mathcal{G}_\theta(\cdot)}{d\theta} \right],
\label{eq2}
\end{equation}
where $\mathcal{D}$ is the discrimator forward process. The score estimator of teacher $\mathcal{T}_\phi(\cdot)$ is itself, the generator's score estimator $\mu_{\text{gen}}^{\psi}(\cdot)$ is initialized with $\mathcal{T}_\phi(\cdot)$, and is dynamiclly updated to track $p_{\text{gen}}$ with diffusion loss:
\begin{equation}
\mathcal{L}_{\text{Diffusion}} = \mathbb{E}_{x_{t-1},t,\epsilon\sim\mathcal{N}(\mathbf{0},\mathbf{I})} [\|\mu^\psi_{\text{gen}}(x_t, t) - \epsilon\|_2^2],
\label{eq3}
\end{equation}
DMD2 reuses the parameter $\psi$ of $\mu_{\text{gen}}^\psi(x_t,t)$ and extra trainable heads to distinguish $p_{\text{gen}}$ and real image distribution $p_{\text{real}}$, gradient of their adversarial objective w.r.t. $\psi$ is,
\begin{equation}
\begin{split}
\mathcal{P}\left(\cdot\right) &= \log\mathcal{D}\left(\cdot\right)\frac{dD\left(\cdot\right)}{d\psi}, \\
\nabla_\psi \mathcal{L}_{\text{AdvDisc}} &= \mathbb{E}_{z, t, x\sim p_{\text{real}}} \left[\mathcal{P}\left(x\right) - \mathcal{P}\left(\mathcal{G}_\theta(\cdot)\right) \right],
\end{split}
\label{eq4}
\end{equation}

\subsection{Training Inefficiency of DMD Series}
Despite their impressive performance, methods in the DMD series are characterized by significant computational overhead during distillation. This is evident in the extensive training schedules required by prominent models. For instance, the original DMD\citep{DMD} required $20,000$ iterations with a batch size of 2,304 to distill Stable Diffusion v1.5 \citep{latentdiffusion} for single-step generation. Similarly, DMD2 \citep{DMD2} used $24,000$ iterations to distill SDXL \citep{sdxl} for four-step generation, and ADM \citep{ADM} used $16,000$ iterations for single-step SDXL distillation. Given the strong empirical results and open-source implementation of DMD2, we select it as the foundation for our investigation into these inefficiencies. One primary source of inefficiency in DMD2 stems from its optimization strategy. As noted by \cite{cheng2025pose}, DMD2 simultaneously optimizes the model using two distinct gradients: a distribution-matching gradient (Eq.~\eqref{eq1}) and an adversarial gradient (Eq.~\eqref{eq2}). A direct summation of these gradients can introduce conflicting objectives, potentially steering the model toward a suboptimal state. This conflict can degrade both the accuracy of the distribution matching and the perceptual quality of the generated images, thereby hindering efficient convergence.
A second challenge lies in the dual role assigned to the generator's score estimator. It is tasked with two demanding objectives: tracking the output distribution of $\mathcal{G}_{\theta}(\cdot)$(Eq.~\eqref{eq3}) and discriminating between real and generated samples (Eq.~\eqref{eq4}). To stabilize this complex dynamic, a two-time scale update rule (TTUR) is employed in DMD2, where score estimator is updated five times for every single update of the generator $\mathcal{G}_{\theta}(\cdot)$. This significantly contributes to the model's overall training inefficiency.

\subsection{Faster Convergence at First Stage}
\label{sec:phase1}
\paragraph{Adversarial Training is Necessary.} DMD2 framework optimizes the generator by naively summing the Distribution-Matching (DM) loss from the teacher and an adversarial loss against real images at every timestep. This superposition of gradients can result in suboptimal and inefficient optimization. When we remove the adversarial teacher entirely, we observe that under pure DM loss supervision, the generator rapidly converges to a suboptimal domain, producing outputs with unnaturally high contrast and lacking fine-grained textures. We attribute this behavior to the mode-seeking nature of the reverse KL divergence, an observation also discussed in ADM \citep{ADM}. This finding underscores the necessity of the adversarial loss with real images for perceptual fidelity.
\paragraph{Decoupling Losses with a Timestep-Aware Strategy.}
We observe that the generator's objective changes throughout the denoising process. For a Few-step distilled model, the initial, high-noise timesteps (low Signal-to-Noise Ratio, or SNR) primarily establish global composition and structure, and the low-noise timesteps (high SNR) focus on refining details, textures, and color tones to enhance realism. This observation is corroborated by findings of \cite{cheng2025pose} in video generation tasks, which noted that the adversarial training in DMD2 is most active at high SNRs, whereas DM loss excels at guiding the model through high-noise regimes. Based on these insights, we assign DM loss and adversarial loss to distinct timesteps: 
\textbf{\textit{1.During the high-noise regime}}, we optimize the generator exclusively with the DM loss (Eq.~\eqref{eq1}). This allows the model to efficiently learn the teacher's fundamental distribution and ODE trajectory in the early phases of generation. 
\textbf{\textit{2.For the low-noise step}}, we apply the adversarial loss against real images, enabling the model to refine perceptual quality and texture in the final denoising step. During each generator update, we sample one timestep $t$ from high-noise timesteps and $x_t$ from DMD2’s back-simulation forward process $\mathcal{B}$ to compute the DM loss $\nabla_\theta \mathcal{L}_{\text{DMD}}^{\text{AT}}$
, then employ $\mathcal{B}$ to propagate the denoised output $x_{t-1}$ to a final clean image $x_0$:
\begin{equation}
    x_{t_{1}}=\mathcal{G}_{\theta}(x_t,t);\space x_0= \texttt{Detach}(\mathcal{B}(x_{t_1},0)),
\end{equation}
where $\texttt{Detach}$ denotes stop gradient, 
$x_0$ is then used for the adversarial loss computation. We perform diffusion forward on $x_0$ at $\hat{t}$ from low-noise timesteps to obtain the noisy sample $\hat{x}$. Gradient for adversarial loss is:
\begin{equation}
\nabla_\theta \mathcal{L}_{\text{AdvGen}}^{\text{TA}} = \left[\mathbb{E}_{\hat{t},\hat{x}} \log\mathcal{D}\left(\mathcal{V}\left(\mathcal{G}_\theta(\hat{x}, \hat{t})\right) \right)\frac{d\mathcal{G}_\theta(\cdot)}{d\theta} \right],
\label{eq:timestep_aware_gen}
\end{equation}
where the $\mathcal{D}(\cdot)$ is the pixel-level discriminator, and $\mathcal{V}$ is the decode process of VAE. This timestep-aware strategy reduces interference between these two optimization objectives. Our experiments prove that this approach significantly improves training efficiency while generating high-quality images with enhanced realism and textural detail.
\vspace{-4mm}
\paragraph{Pixel-GAN Alleviates Mode-Seeking.}
To enforce realism and structural coherence, the present study performs adversarial learning directly in the pixel space utilizing a discriminator. In contrast to a conventional latent-space GAN, the discriminator in our model is constructed upon the frozen vision encoder of the Segment Anything Model (SAM)\cite{sam} to extract hierarchical features with multiple trainable discriminator heads attached. The discriminator's trainable parameters $\omega$ are updated via:
\begin{equation}
\mathcal{L}_{\text{AdvDisc}}^{\text{PG}} = \mathbb{E}_{x_{\text{real}}} \left[-\log\mathcal{D}_\omega\left(\cdot\right)\right] + \mathbb{E}_{z} \left[ \log\mathcal{D}_\omega\left(\mathcal{V}(\cdot)\right) \right],
\label{eq:pixelgan_disc}
\end{equation}
The discriminator is characterized by its exceptional sensitivity to local geometric structures and fine-grained textures, a capability that is facilitated by SAM's powerful, general-purpose representations, as noted by \cite{ADM}. This pixel-level supervision exerts a stringent realism constraint from the training's earliest stages, compelling the generator to expeditiously discern and anchor to diverse, high-fidelity modes within the data distribution. Visualizations and experiments prove it effectively prevents premature convergence to simplistic or blurry solutions (mode-seeking).
\vspace{-4mm}
\paragraph{Stabilize Score Estimator.}
In contrast to DMD2~\cite{DMD2}, where the score estimator is required to serve as a discriminator (Eq.~\eqref{eq4}) and thus faces conflicting optimization objectives, our approach trains $\mu_{\text{gen}}^\psi$ solely via the diffusion loss (Eq.~\eqref{eq3}), eliminating the tasking burden and training complexity for $\mathcal{G}_{\theta}(\cdot)$. Our experiments show that updating the score estimator only once or twice per generator update (TTUR=1,2) is sufficient for stable and accurate distribution tracking. This lightweight coupling leads to more stable training dynamics and superior sample fidelity compared to DMD2 with TTUR=5, while reducing computational overhead. Similar to implicit distribution alignment proposed by \cite{senseflow}, we also adopt an Exponential Moving Average (EMA) update strategy to ensure that the score estimator $\mu_{\text{gen}}^\psi$ accurately tracks the evolving distribution of the generator $\mathcal{G}_{\theta}(\cdot)$. Specifically, after each generator update, we inject the latest generator parameters into the score estimator using an EMA coefficient $\lambda_{\text{ema}}$, i.e.,
\begin{equation}
    \psi \leftarrow \lambda_{\text{ema}} \psi + (1 - \lambda_{\text{ema}}) \theta,
\end{equation}
which enables $\mu_{\text{gen}}^\psi$ to closely follow the generator's trajectory with minimal additional updates. 
\vspace{-4mm}
\paragraph{Putting Everything Together.} We introduce \ours, a highly efficient framework for Distribution Matching Distillation.
In summary, \ours's training objectives via a timestep-aware strategy efficiently distill the fundamental distribution of the teacher model in low-SNR timesteps and refine perceptual quality and texture in the final high-SNR timestep. To counteract the mode-seeking tendency of the DM loss, we introduce a SAM-based Pixel-GAN that robustly enhances realism. The combination of these strategies and the stabilized score estimator enables a more effective and balanced optimization of $\mathcal{G}_{\theta}(\cdot)$.

\subsection{Reinforcement Learning for Distilled Model} 

Using the training paradigm above, we have developed a student generator that can compete with the teacher model. Subsequently, our focus shifts to enhancing its performance beyond that of the teacher and deploying it in practical scenarios. Preference optimization provides a direct way to improve image fidelity and detail richness effectively on diffusion models. However, previous attempts on few-step reinforcement learning like PSO\cite{pso} and HyperSD\cite{hypersd} encounter serious reward hacking, a phenomenon overfitting on oil painting or smoothed images with less details.
\vspace{-4mm}
\paragraph{Reasons for Serious Reward Hacking.}
PSO\cite{pso} and HyperSD\cite{hypersd} rely on the sampling trajectory to denoise noisy latent iteratively. Preference optimization on clean images confines gradient backpropagation to low-noise timesteps, making the model overfit the reward biases, where the model prioritizes superficial features (e.g., specific color palettes). Specifically, HyperSD utilizes ImageReward \citep{xu2023imagereward} and produces overexposed and oil-painted results, as shown in Fig.\ref{fig:performance_cmp_rl}. PSO selects PickScore\citep{pickscore} as a reward model and generates smoothed images.
\paragraph{Improved Preference Optimization for Distilled Model.}
The direct solution is to cover the sampling trajectory, including high-noise timesteps. First, reward models are required to score noisy latent representations at any timestep. LRM\cite{zhang2025diffusion} inherently meets our demands. Multiple candidates are sampled at each timestep from shared noise initialization and are rated by LRM to construct win-lose pairs. We further find that not all timesteps are necessary. As shown in Fig.\ref{fig:sample_test}, with the same initial noise, it is evident that images sampled by few-step distilled model at high-noise steps exhibit better diversity in layout and fine-grained details compared to those from low-noise steps. As a result, we only perform stochastic sampling in the high-noise phase, altering latent representations that are deemed optimal/suboptimal by the reward model. Second, we combine the logarithmic likelihood loss with the vanilla loss of \ours~ during the training process rather than applying preference optimization separately for stable training. We show that the combination can further boost the generation performance of the accelerated diffusion models towards human preference and text-image alignment.

\input{figures/sample_test}

\paragraph{Formal Descriptions.}
Given a generator ${\mathcal{G}_\theta(\cdot)}$ distilled from a pretrained diffusion model $\mathcal{T}_{\phi}(\cdot)$, it can sample clean images from pure noise $z_T\sim \mathcal{N}(\mathbf{0}, \mathbf{I})$, conditioned on text prompt $c$, within $T=4$ steps. At high-noise timesteps, we sample a set of $k$ noisy latent images $\{z_{t-1}^1,...,z_{t-1}^k\}$ from the same initial latent image $z_t$. LRM predicts preference scores. The samples corresponding to the highest and lowest normalized scores are selected as win-lose pairs, thereby constructing paired training data $(z_{t},z_{t-1}^w,z_{t-1}^l)$ to the sampling pool. These pairs are subsequently used to minimize the loss function:
\begin{align}
\mathcal{L}_{rl} &= -\mathbb{E} \left[ \log \sigma \left( \beta \mathcal{H}(w,l) \right) \right], \\
\mathcal{H}(w,l) &= \log \frac{p_\theta(z_{t-1}^w|z_t,c)}{p_{ref}(z_{t-1}^w|z_t,c)} - \log \frac{p_\theta(z_{t-1}^l|z_t,c)}{p_{ref}(z_{t-1}^l|z_t,c)},
\label{lpo}
\end{align}
where $\sigma$ and $\beta$ are inherent regularization constants, $p_*({z_{t-1}^*|z_t, c)}$ denotes the backward process to denoise $z_t$ in the LCM scheduler.

%% file: figures/sample_test.tex
\begin{figure}[htbp]
    \centering 
    \includegraphics[width=0.5\textwidth, angle=0]{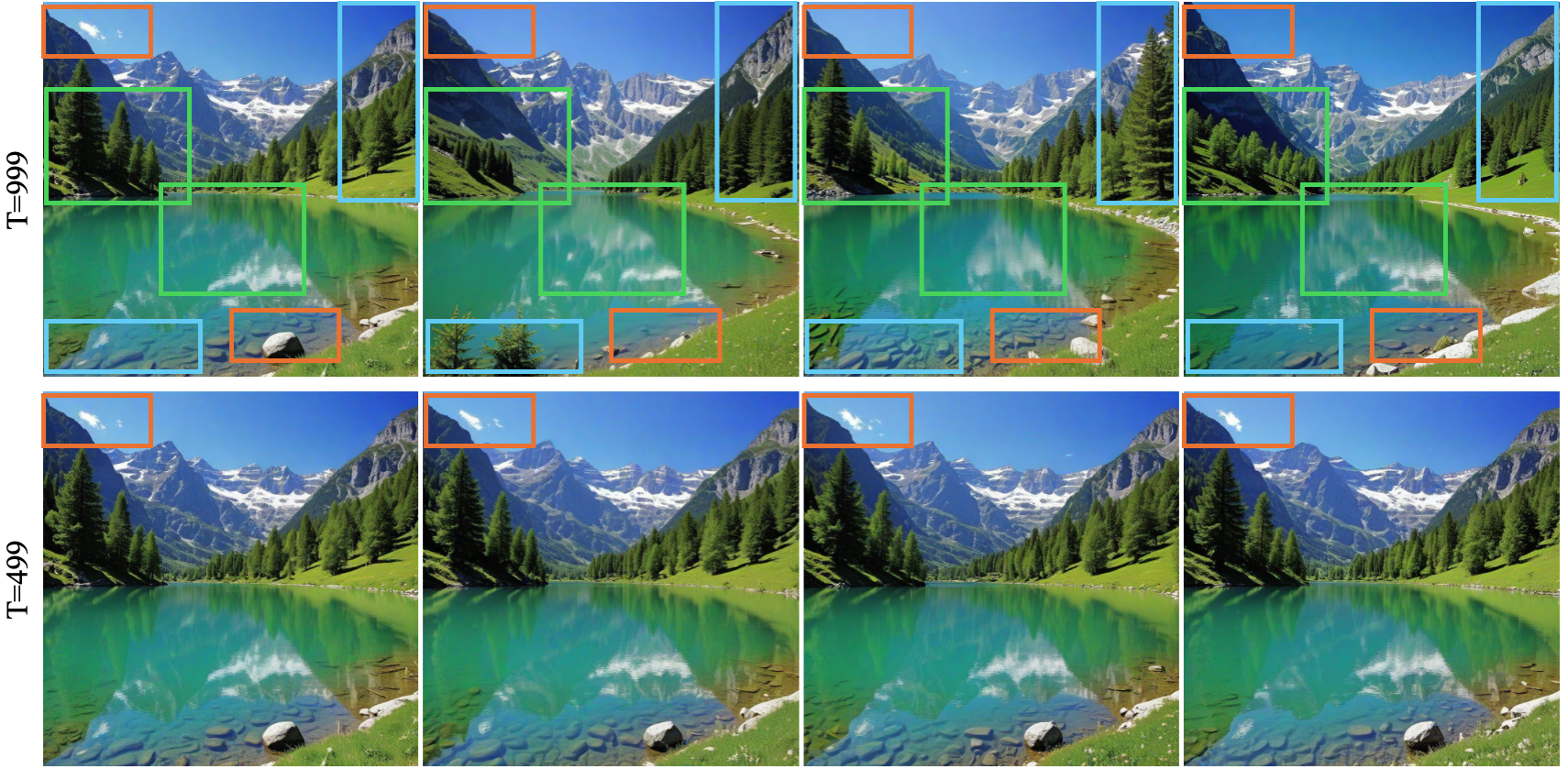}
    \vspace{-6mm}
    \caption{Sampling variance analysis at different time steps. The first row displays samples obtained at the 999th denoising step, while the second row corresponds to the 499th step.}
    \label{fig:sample_test}
    \vspace{-4mm}
\end{figure}

%% file: sec/4_exp.tex
\vspace{-2mm}
\section{Experiments}
\subsection{Implementation Details}
\paragraph{Experiment Setup}
For the first phase, we conduct experiment on Score-based diffusion model SDXL\cite{sdxl} and Flow Matching based SD3-Medium\cite{sd3}. We utilize a filtered set from the LAION 5B \citep{schuhmann2022laion} dataset to provide high-quality image-text pairs for training, following the setting of \citep{yin2024improved}. For discriminator conditioning, we adopt the vision encoder from \cite{sam} as the backbone to extract image representations. The structure of trainable discriminator heads follows the 2D architecture of \citep{ADM}. For the second phase, we adopt the training dataset from the first phase and utilize the Latent Reward Model from \citep{lpo}, and experiment on distilled 4-step SDXL. We sample a set of noisy latent images at the high-noise timestep $t={749,999}$ and set $k=4$. We conduct experiments on NVIDIA H20 GPUs.
\vspace{-2mm}
\paragraph{Evaluation Tasks and Baseline}
The evaluation of image generators is conducted on 10K prompts from COCO 2014~\citep{lin2014microsoft}, adhering to the DMD2~\citep{yin2024improved} framework, containing 10,000 images. We present the result of CLIP score~\citep{radford2021learning} (ViT-B/32) to evaluate text-image similarity, and we adopt a set of advanced preference-based metrics to thoroughly evaluate the quality of generated images from multiple human-aligned perspectives. 
We use HPSv2~\citep{hpsv2} to measure fine-grained image-text semantic alignment, focusing on how well the generated content adheres to the input prompt. 
ImageReward~\citep{xu2023imagereward} and PickScore~\citep{pickscore} are employed to assess overall aesthetic quality and perceptual appeal, reflecting general human preferences in visual coherence and composition. Furthermore, we evaluate multidimensional human preferences using MPS~\citep{MPS}, a recently proposed metric that captures diverse aspects of human judgment, such as object accuracy, spatial relation, and attribute binding-beyond global similarity. Together, these metrics provide a comprehensive and human-centric evaluation of both fidelity and preference in text-to-image generation.
To demonstrate the effectiveness of phase 1 distillation, we compare our 4-step generative models against SDXL\citep{sdxl}, as well as other open-sourced timestep distillation methods, including LCM-SDXL~\citep{LCM}, SD3-Lighting\cite{sdxllighting}, SDXL-Turbo~\citep{ADD}, Realism and Vibrant version of NitroSD~\cite{nitro}, Flash-SD3\cite{chadebec2025flash}, and DMD2~\citep{DMD2}. For phase 2, we further evaluate our approach by comparing it with three reinforcement learning-finetuned models, Hyper-SDXL~\citep{hypersd}, PSO-DMD2~\cite{pso}, and LPO-SDXL~\cite{lpo}.

\subsection{Experiment Analysis}
\input{tables/main_compare}
\input{tables/sd3_cmp}
\input{tables/rl_results}
\paragraph{Phase 1: Efficient Distillation}
Our method achieves highly efficient distillation from the teacher model, leading to state-of-the-art performance across all benchmarks, by decoupling the distribution matching (DM) and adversarial losses with a timestep-aware strategy, employing PixGAN to alleviate mode-seeking behavior, and stabilizing the generator's score estimator. 
As shown in Tab.~\ref{tab:phase1_main}, we distill SDXL using various two-time update rules (TTUR), which corresponds to the update frequency ratio between the score estimator and generator. In contrast to DMD2, which uses a TTUR of 5 and thus hinders training efficiency, we experiment with TTUR values of 1, 2, and 5. Our results demonstrate significant improvements in both efficiency and performance. With a TTUR of 5, our model surpasses DMD2 on all benchmarks while requiring only $37.5\%$ of the training cost (batch size $\times$ training steps). Reducing the TTUR to 2 allows us to maintain superior human preference scores and comparable text-image consistency with only $8.3\%$ of DMD2's training cost. In the most extreme case, training for only 1,000 steps with a TTUR of 1, with merely $2.1\%$ of DMD2's training cost, we still yield a higher human preference score. Notably, under all tested settings, our model consistently outperforms the original teacher model. We also extend \ours~ to SD3-Medium with solid performance, outperforming other methods. Results are shown in Tab.~\ref{tab:phase1_sd3}. This further demonstrates the generalizability of our method. We further extend \ours~to the SD3-Medium\cite{sd3} model under the Flow Matching framework, employing LoRA, with a TTUR ratio of 2. As shown in Tab.~\ref{tab:phase1_sd3}, our approach achieves significantly better results using only 4K training steps, substantially outperforming both the teacher model (NFE=28, CFG=7) and SD3-Flash\cite{chadebec2025flash} (NFE=4, CFG=0). This demonstrates that our distillation strategy is also highly effective and efficient within the Flow Matching paradigm.
\vspace{-2mm}
\paragraph{Phase 2: Boost Performance with RL.}
Incorporating reinforcement learning, \ours~achieves a performance comparable to other reinforcement approaches with fewer computational resources, as shown in the Tab. \ref{tab:rl_results} and Fig. \ref{fig:performance_cmp_rl}. \ours~scores the highest on PickScore and MPS, and image fidelity surpasses SDXL and other competitors. Although Hyper-SDXL has the highest ImageReward score and HPSv2 score, it generates overexposed colors and unnatural images. LPO-SDXL gets the highest CLIP score, but it produces oversmoothed images. We speculate that the reason may be that these models use the trained models directly for reinforcement training. This is prone to reward hacking and only rewards the results preferred by the reward model. Meanwhile, \ours~introduces preference optimization during the training process of Phase 1, with the constraints from distribution matching and PixelGAN, which alleviates the problem of reward hacking.

\input{figures/performance_cmp_rl}

\subsection{Ablation Studies}
\input{figures/ttur2_cmp}
\paragraph{Comparision with DMD2 under Samller TTUR.}
We set TTUR=2 for ablate the performance of DMD2 and \ours~under phase 1. The results, presented in Fig. \ref{fig:ttur2_cmp}, show that our method exhibits stable and continuous improvement throughout the training process. In contrast, DMD2 shows slight initial gains but quickly degrades as training progresses. This comparison validates that our approach offers much better training stability and efficiency.

\input{figures/ema_cmp}
\vspace{-3mm}
\paragraph{Phase 1: Significance of EMA in Score Estimator.}
As described in Sec.\ref{sec:phase1}, we employ an Exponential Moving Average (EMA) strategy to help the score estimator more accurately track the generator's distribution, especially under high-frequency updates. To validate the effectiveness of this approach, we conduct an ablation study comparing performance with and without the EMA strategy. As shown in Fig. \ref{fig:ema_cmp}, the model with EMA achieves higher ImageReward and Pickapic scores in later training stages. The HPSv2 scores remain nearly identical. This confirms that the EMA strategy enhances visual quality and human preference without compromising text alignment.

\input{figures/ablation_gan}
\input{tables/rl_performance}
\vspace{-3mm}
\paragraph{Phase 2: Trade-Off in Time Scale Update Rule for RL.}
Rather than superimposing multiple loss functions via a weighted sum, we use an alternating update strategy to update the generator, applying different loss functions at different frequencies. We initialize the generator and the fake score estimator with weights from TTUR1-1k experiment, and the real score estimator is initialized with SDXL. 
We train the model for 2,000 iterations on a single H20 GPU under different frequency ratios between reinforcement loss and distribution matching loss (1:1, 2:1, 5:1, 10:1). The metric comparisons are shown in Tab. \ref{tab:rl_performance}. The 5:1 ratio achieves the highest score among these settings.

\vspace{-3mm}
\paragraph{Phase 2: Other Ablation Experiments on Reinforcement Learning.}
\textit{1. Online training VS Post-training:} We compare online training with post-training by LPO \citep{zhang2025diffusion} alone. Our method demonstrates superior performance over Post-Train LPO, which validates the advantage of our proposed training paradigm.
\textit{2.High-noise VS all noise:} Training only on high-noise steps achieves better results than training on all-noise steps. 
\textit{3. Including PixelGAN loss:} Furthermore, the incorporation of an additional Pixel-Gan objective yields a positive gain, resulting in a marginal improvement in the metrics. Fig.\ref{fig:ablation_gan} supplements the results with and without GAN loss across 1,000 to 5,000 iterations. We selected the 5k-step model with GAN loss as our final enhanced version, as it delivered the best performance. The training cost for this model was 12 H20 GPU hours.


%% file: tables/main_compare.tex
\newcommand{\updownarrows}[1]{\raisebox{-.2ex}{\scalebox{1.2}[0.8]{#1}}}
\definecolor{lightgray}{gray}{0.9}
\definecolor{lightblue}{RGB}{230, 240, 255}
\begin{table}[t]
\centering
\setlength{\tabcolsep}{2pt}
\caption{Comparison of \ours~on SDXL under stage 1 with other distillation methods on the COCO-10k dataset. \textbf{ImgRwd} denotes ImageReward score. \textbf{Cost} refers to the product of batch size and training iterations. Best performance is highlight with \textbf{Bold}, and the second is with \underline{underline}.}
\vspace{-2mm}
\label{tab:phase1_main}
\resizebox{0.5\textwidth}{!}{
\begin{tabular}{lccccccc}
\toprule
\textbf{Method} & \textbf{\#NFE} & \textbf{ImgRwd} \updownarrows{$\uparrow$} & \textbf{CLIP} \updownarrows{$\uparrow$} & \textbf{Pick} \updownarrows{$\uparrow$} & \textbf{HPSv2} \updownarrows{$\uparrow$} & \textbf{MPS} \updownarrows{$\uparrow$} & \textbf{Cost} \updownarrows{$\downarrow$} \\
\midrule
\rowcolor{lightgray}
SDXL & 100 & 0.7143 & 0.3295 & 0.2265 & 0.2865 & 11.87 & - \\
\midrule
LCM-SDXL & 4 & 0.5562 & 0.3250 & 0.2236 & 0.2818 & 11.11 & - \\
SDXL-Lightning & 4 & 0.6952 & 0.3268 & 0.2285 & 0.2888 & 12.15 & - \\
SDXL-Turbo & 4 & 0.8338 & 0.3302 & 0.2286 & 0.2899 & 12.25 & - \\
NitroSD-Realism & 4 & 0.9112 & 0.3274 & 0.2291 & 0.2975 & 12.43 & - \\
NitroSD-Vibrant & 4 & 0.8419 & 0.3201 & 0.2205 & 0.2865 & 11.13 & - \\
DMD2-SDXL & 4 & 0.8748 & \textbf{0.3302} & 0.2309 & 0.2937 & 12.41 & 128*24k \\
\midrule
\multicolumn{8}{c}{\textbf{\ours~under Phase 1}} \\
\midrule
\rowcolor{lightblue}
TTUR1-1k & 4 & \underline{0.9509} & 0.3292 & \underline{0.2322} & 0.2968 & \underline{12.67} & \textbf{64k (2.1\%)} \\
TTUR2-4k & 4 & 0.9450 & 0.3291 & \underline{0.2322} & 0.2969 & 12.65 & \underline{64*4k} \\
TTUR2-8k & 4 & \textbf{0.9740} & \underline{0.3298} & \textbf{0.2327} & \underline{0.2981} & \textbf{12.71} & 64*8k \\
TTUR5-18k & 4	& 0.9426  &	\textbf{0.3302} & 0.2319 & \textbf{0.2982} & 12.63 & 64*18k\\
\bottomrule
\end{tabular}
}
\end{table}

%% file: tables/sd3_cmp.tex
\definecolor{lightgray}{gray}{0.9}
\definecolor{lightblue}{RGB}{230, 240, 255}
\begin{table}[t]
\centering
\setlength{\tabcolsep}{4pt}
\caption{Comparison of \ours~on SD3 under stage 1 with other distillation method and baseline on COCO-10k dataset.}
\label{tab:phase1_sd3}
\resizebox{0.5\textwidth}{!}{
\begin{tabular}{lccccccc}
\toprule
\textbf{Method} & \textbf{\#NFE} & \textbf{ImgRwd} \updownarrows{$\uparrow$} & \textbf{CLIP} \updownarrows{$\uparrow$} & \textbf{Pick} \updownarrows{$\uparrow$} & \textbf{HPSv2} \updownarrows{$\uparrow$} & \textbf{MPS} \updownarrows{$\uparrow$} & \textbf{Cost} \updownarrows{$\downarrow$} \\
\midrule
\rowcolor{lightgray}
SD3-Medium & 28 & 1.0173 & 0.3301 & 0.2273 & 0.2933 & 12.05 & - \\
\midrule
SD3-Flash & 4 & 0.8459 & 0.3258 & 0.2259 & 0.2849 & 11.83 & - \\
\midrule
\multicolumn{8}{c}{\textbf{\ours~under Phase 1}} \\
\midrule
\rowcolor{lightblue}
TTUR2-4k & 4 & \underline{1.0193} & \underline{0.3269} & \underline{0.2285} & \textbf{0.2976} & \underline{12.43} & \textbf{32*4k} \\
TTUR2-7k & 4 & \textbf{1.0214} & \textbf{0.3266} & \textbf{0.2286} & \underline{0.2975} & \textbf{12.46} & \underline{32*7k} \\
\bottomrule
\end{tabular}
}
\end{table}

%% file: tables/rl_results.tex
\definecolor{lightblue}{RGB}{230, 240, 255}
\begin{table}[t]
\centering
\setlength{\tabcolsep}{3pt}
\caption{Comparison of \ours~under phase2 with other models with reinforcement learning on COCO-10k dataset.}
\vspace{-2mm}
\label{tab:rl_results}
\resizebox{0.5\textwidth}{!}{
\begin{tabular}{lccccccc}
\toprule
\textbf{Method} & \textbf{\#NFE} & \textbf{ImgRwd} \updownarrows{$\uparrow$} & \textbf{CLIP} \updownarrows{$\uparrow$} & \textbf{Pick} \updownarrows{$\uparrow$} & \textbf{HPSv2} \updownarrows{$\uparrow$} & \textbf{MPS} \updownarrows{$\uparrow$} & \textbf{GPU Hours}\\
\midrule
Hyper-SDXL & 4 & \textbf{1.085} & \underline{0.3300} & 0.2324 & \textbf{0.3030} & 12.45 & 400 A100 \\
PSO-DMD2 & 4 & 0.9157 & 0.3285 & 0.2338 & 0.2897 & 12.53 & 160 A100 \\ 
LPO-SDXL & 40 & \underline{1.0417} & \textbf{0.3324} & \underline{0.2342} & \underline{0.2965} & \underline{12.58} & \underline{92 A100} \\
\rowcolor{lightblue}
\ours & 4 & 1.0035 & 0.3285 & \textbf{0.2346} & 0.2930 & \textbf{12.84} & \textbf{12 H20} \\
\bottomrule
\end{tabular}
}
\end{table}

%% file: figures/performance_cmp_rl.tex
\begin{figure}[t]
\centering
\includegraphics[width=1.1\linewidth]{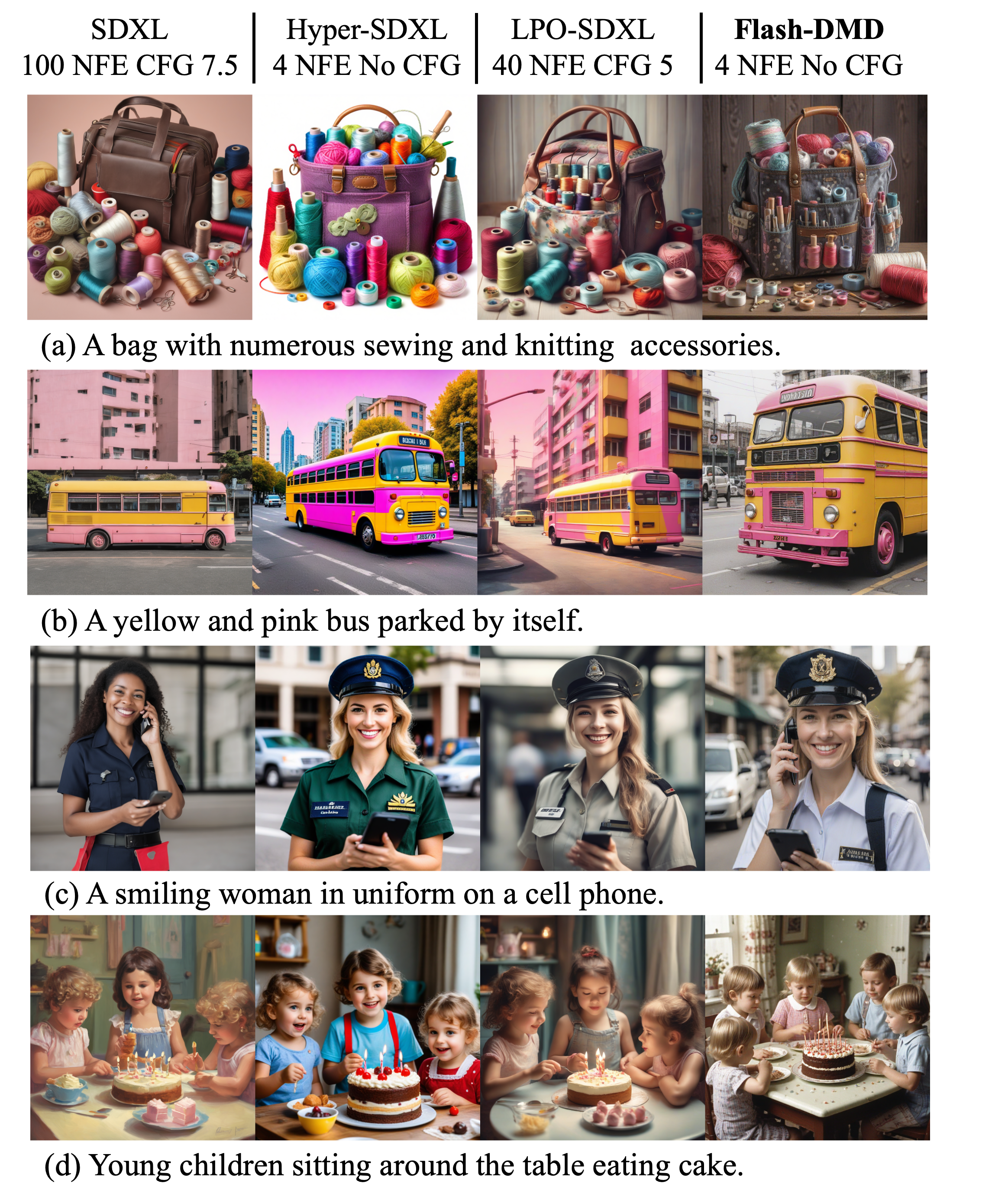}
\vspace{-6mm}
\caption{
   Qualitative comparisons with other reinforcement approaches on SDXL.
com}
\vspace{-2mm}
\label{fig:performance_cmp_rl}
\end{figure}

%% file: figures/ttur2_cmp.tex
\begin{figure}[t]
\centering
\includegraphics[width=\linewidth]{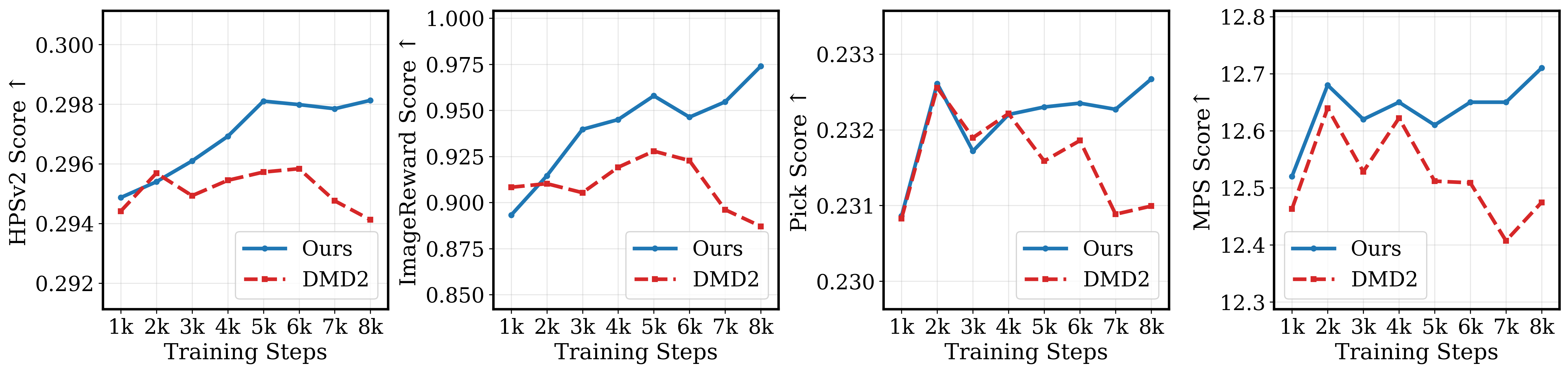}
\vspace{-4mm}
\caption{
    Evaluation results of DMD2(red) and \ours~(blue) with TTUR at the ratio of 2 on SDXL. 
}
\vspace{-2mm}
\label{fig:ttur2_cmp}
\end{figure}

%% file: figures/ema_cmp.tex
\begin{figure}[t]
\centering
\includegraphics[width=\linewidth]{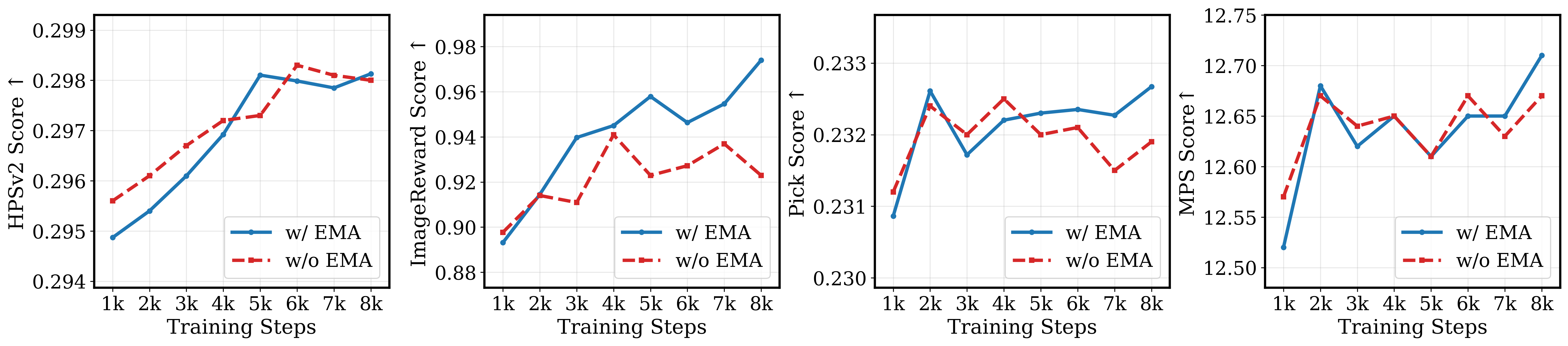}
\caption{
    Evaluation results of \ours~(ours) with or without EMA on ImageReward, PickScore, and HPSv2. The training steps range from 1,000 to 8,000. Both models are trained with a two-time scale update rule (TTUR). The generator and the score estimator are updated at a rate of 1:2, i.e., TTUR=2.
}
\label{fig:ema_cmp}
\end{figure}

%% file: figures/ablation_gan.tex
\begin{figure}[t]
    \centering 
    \includegraphics[width=0.5\textwidth, angle=0]{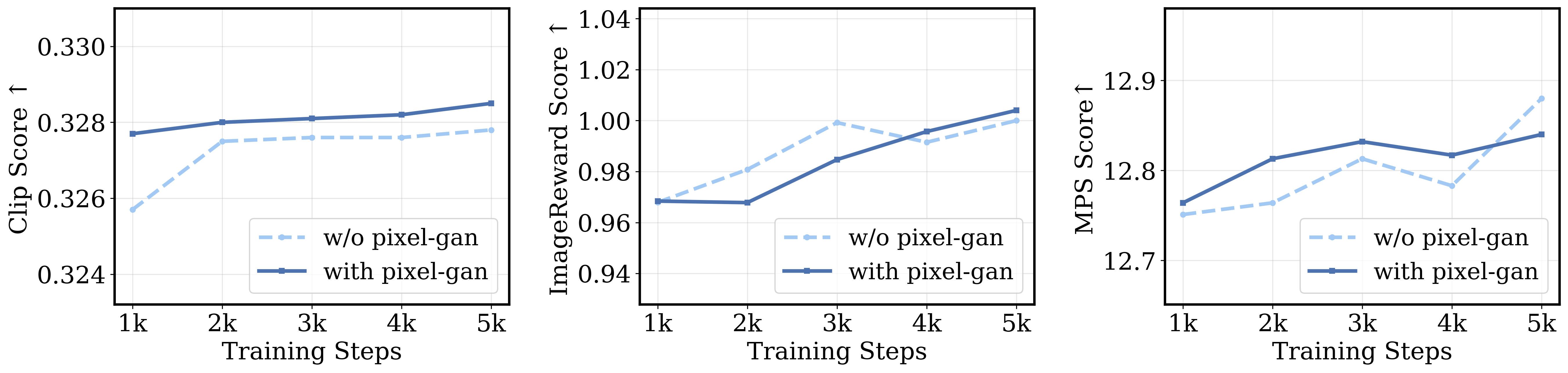}
    \caption{Evaluation results of Reinforcement Learning with and without pixel-GAN. Both models use 5:1 setting.}
    \vspace{-5mm}
    \label{fig:ablation_gan}
\end{figure}

%% file: tables/rl_performance.tex
\begin{table}[t]
\centering
\setlength{\tabcolsep}{2pt}
\scriptsize
\caption{Comparison of \ours~at stage2 with different variants in reinforcement learning experiments .}
\label{tab:rl_performance}
\resizebox{0.5\textwidth}{!}{
\begin{tabular}{ccccccc}
\toprule
\textbf{Method} & \textbf{ImgRwd} & \textbf{CLIP} & \textbf{PickScore} & \textbf{HPSv2}& \textbf{MPS} & \textbf{GPU Hours} \\
\midrule
$\text{TTUR1-1k}_{~\text{Phase 1}}$ & 0.9508 & 0.3292 & 0.2322 &$\textbf{0.296}$ & 12.672 & -- \\
\midrule
1:1 & 0.9135 & 0.3284 & 0.2330 &0.2945 & 12.755 & 5.7 \\
2:1 & 0.9315 & 0.3271 & 0.2329 &0.2942 & 12.770 & 5.2 \\
5:1 & 0.9808 & 0.3275 & 0.2345 &0.2904 & 12.764 & 4.8 \\
10:1 & 0.9640 & 0.3272 & 0.2344 &0.2874 & 12.685 & 4.7 \\
\midrule
Post-Train LPO & 0.9795 & 0.3284 & 0.2345 &0.2882 & 12.689 & 5.0 \\
all noise & 0.9421 & 0.3294 & 0.2331 &0.2925 & 12.800 & 7.3 \\
+ pixelgan & 0.9678 & 0.3280 & 0.2345 &0.2914 & 12.812 & 5.8 \\
\rowcolor{lightblue}
$\text{\ours}~_{\text{Phase 2}}$ & $\textbf{1.0004}$ & $\textbf{0.3285}$ & $\textbf{0.2346}$ &0.2931 & $\textbf{12.813}$ & 12.0 \\
\bottomrule
\end{tabular}
}
\end{table}

%% file: sec/5_conclude.tex
\section{Conclusion}
In this paper, we present \ours, a twofold approach that addresses the inefficiencies of existing diffusion distillation methods by via timestep-aware objectives and optimizing the distillation process. In the early phase, \ours~accelerates convergence by coordinating distribution matching and perceptual realism enhancement. In the later phase, it refines visual details using latent reinforcement learning while preventing mode collapse and artifacts. Experiments show \ours~achieves superior generation quality and the highest human preference scores with significantly reduced training costs. Our method makes diffusion distillation more efficient and accessible, paving the way for advancements in low-step generative modeling.

%% file: sec/X_suppl.tex
\clearpage
\setcounter{page}{1}
\maketitlesupplementary

\begin{figure*}[t]
\centering
\includegraphics[width=\linewidth]{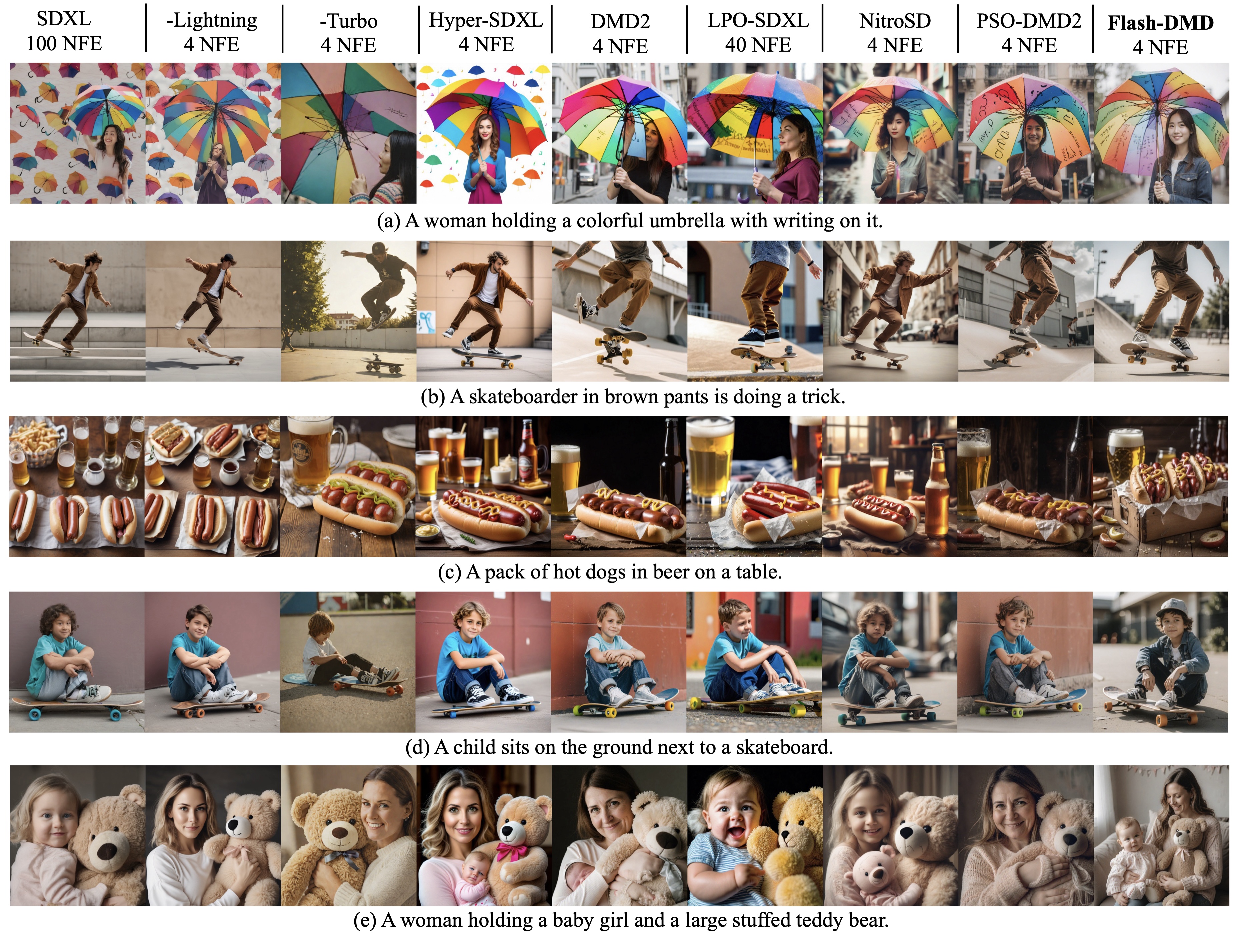}
\vspace{-5mm}
\caption{
   Qualitative comparisons with other models.
}
\label{fig:comp_models_v1}
\end{figure*}

\input{algorithm/stage1}

\section{Latent Reward Model: Selection Rationale}
As noted earlier, the previous practice in online preference optimization \cite{pso, hypersd} involves applying reward models\cite{liao2025step,pickscore,xu2023imagereward,MPS,hpsv2,spo,visionreward} in the final denoising stage, leading to reward hacking. A natural question arises: \textit{why not deploy optimization across a more diverse set of timesteps? }
\noindent
Two primary factors explain this limitation. Computationally, reward evaluation would necessitate VAE decoding at multiple steps, which would increase GPU memory and computation requirements. More fundamentally, their reward models lack timestep sensitivity, and the training paradigms are time-agnostic, having been developed on static image datasets without diffusion process context.
\noindent
We empirically investigate the prevalent reward models and systematically summarize their characteristics in the Tab.~\ref{tab:reward_models}, including scope space, evaluation dimensions, and consideration of timestep. In light of the analysis above, LRM \cite{lpo} is the most suitable choice. It can evaluate noisy latent at any timestep without switching to pixel space. Specifically, it takes a noisy latent and feeds it into a pretrained diffusion model, leveraging the model's native understanding of latent representation across all noise levels.

\definecolor{lightgray}{gray}{0.9}
\definecolor{lightblue}{RGB}{230, 240, 255}
\begin{table}[t]
\centering
\setlength{\tabcolsep}{4pt}
\caption{Comparison of mainstream reward models in four aspects: Scope, Evaluation Dimensions, utilized Feature Extractor (FE) and Timestep Awareness. \textbf{EvalDim} denotes evaluation dimension, \textbf{Time-aware} denotes timestep awareness. \textbf{Aes} denotes aesthetics, \textbf{Align} denotes text-image alignment, \textbf{Fid} denotes fidelity.}
\vspace{-2mm}
\label{tab:reward_models}
\resizebox{0.48\textwidth}{!}{
\begin{tabular}{lcccc}
\toprule
\textbf{Method} & \textbf{EvalDim}  & \textbf{Space} & \textbf{Time-aware} & \textbf{FE} \\
\midrule
PickScore \cite{pickscore} & Pixel &Fid & $\times$ & CLIP \cite{clip}\\
ImageReward \cite{xu2023imagereward} & Pixel &Aes+Fid  & $\times$ & BLIP \cite{li2022blip}\\
MPS \cite{MPS} & Pixel  & Aes+Align+Fid  &$\times$ & CLIP \\
HPSv2 \cite{hpsv2} & Pixel & Align+Fid &$\times$ & CLIP \\
SPM \cite{spo} & Pixel &Aes+Align &$\surd$ & CLIP \\
\rowcolor{lightblue}
LRM \cite{lpo} & Latent &Aes+Align &$\surd$ & SDXL/SD1.5 \cite{sdxl}\\
VisionReward \cite{visionreward} & Pixel &Align+Fid+Safety &$\times$ &Qwen2.5-VL \cite{bai2025qwen2}\\
\bottomrule
\end{tabular}
}
\end{table}


\section{Algorithm of \ours~}
Algorithm \ref{alg:stage1} provides an overview of \ours~. At stage 1, we propose an advanced step distillation method based on distribution matching that converges quickly. Timestep-aware strategy, together with Pixel-GAN constraint, effectively alleviates mode-seeking and improves the realism of generation. At stage 2, we carefully select and incorporate a latent reward model into the training paradigm, further enhancing the image fidelity and image aesthetics.

\section{4-steps and 8-steps \ours~on SDXL}

To demonstrate the effectiveness of \ours~on SDXL~\cite{sdxl}, we extend our method to distill the full SDXL model into an 8-step generative model. In the first stage, we select the timesteps $[999, 874, 749, 629, 499, 374, 249, 124]$ and apply Pixel-GAN at the final timestep. We adopt the Two-Time-Scale Update Rule (TTUR) with a score estimator update frequency of 2, training on the LAION dataset~\cite{schuhmann2022laion} with a batch size of 48. Two variants are trained for 3k and 6k iterations, respectively, denoted as \texttt{TTUR2-3k} and \texttt{TTUR2-6k}.
\noindent
In the second stage, we construct win–lose preference pairs using samples generated from the timesteps $[999, 874]$. We initialize this stage from the \texttt{TTUR2-3k} model (i.e., the 3k-step checkpoint from Stage 1) and continue training for an additional 3k and 6k steps, respectively. The results for Stage 1 and Stage 2 are reported in Tab.~\ref{tab:sdxl8steps} and Tab.~\ref{tab:sdxl8steps_stage2}, respectively. Results show that \ours~keeps outperforming other distillation and reinforcement learning methods at 8-steps generative task, highlighting the advantage of \ours.
\noindent
In terms of subjective results, Fig. \ref{fig:ttur1} and Fig. \ref{fig:4step-lpo} show the results of the 4-step inference at stages 1 and stage 2 of \ours~ based on SDXL, respectively. Fig. \ref{fig:8step} and Fig. \ref{fig:8step-lpo} show the results of the 8-step inference at stages 1 and 2 of \ours~, respectively. Our \ours~ can generate high-quality results with both realism and aesthetic appeal under a small number of steps.
\noindent
Furthermore, we compare our results with SDXL, SDXL-Lighting\citep{sdxllighting}, SDXL-Turbo\citep{ADD}, Hyper-SDXL\citep{hypersd}, DMD2\citep{DMD2}, LPO \citep{zhang2025diffusion}, Realism version of NitroSD\cite{nitro},  PSO\citep{pso}. Fig. \ref{fig:comp_models_v1} demonstrates that our model not only surpasses other distillation models but also outperforms the teacher model in refining image quality.

\input{tables/sdxl_8steps}
\input{tables/sdxl_8steps_stage2}

\section{4-step \ours~on SD3-Medium}
Instead of training on the LAION dataset \cite{schuhmann2022laion}, we curated a proprietary, high-quality training set of 100,000 instances. It encompasses a diverse range of subjects and scenes, including portraits, architecture, flora and fauna, and food—rendered in a realistic photographic style. Our dataset construction followed a structured methodology. First, we generated a set of clear, detailed captions to fully leverage the representational capacity of the SD3 \cite{sd3} text encoders. These captions were then used to synthesize high-quality images using SD3.5 Large, configured with 28 denoising steps and a CFG scale of 4.5. The final training set was curated through a rigorous manual selection process from the generated candidates. We successfully extend \ours~ in SD3-Medium after stage 1, Fig. \ref{fig:flashdmd-4step-sd3} displays some visual results that highlight the strengths of our algorithm.

\input{visuals/visualization}
\section{Additional Visualizations and Captions}
\paragraph{Fig.~\ref{fig:ttur1} captions from left to right, top to bottom are}:
\begin{enumerate}
    \item A cat next to a window behind cans and bottles.
    \item A bunch of red roses bunched together.
    \item A brown teddy bear standing next to bottles of honey.
    \item A brown lamb looking up as other sheep eat hay in a field.
    \item A brown and white horse walking down a road.
    \item A brass-colored vase with a flower bouquet in it.
    \item A boy closely examining a frog in his yard.
    \item A boy in green shorts and a tie posing in front of a tower.
    \item A furry kitten lying on a laptop.
    \item A confection of cake, whipped cream, strawberries, and two candles.
    \item Two birds standing around a box of birdseed.
    \item A soldier riding a red motorcycle down a busy street.
\end{enumerate}

\noindent
\paragraph{Fig.~\ref{fig:ttur2} captions from left to right, top to bottom are:}
\begin{enumerate}
    \item A dog looking up and running to catch a frisbee.
    \item A cake designed to resemble a cup.
    \item A beautiful red-haired woman holding a cup while wearing a sweater.
    \item A row of wooden park benches sitting next to a lake.
    \item A close-up of a baseball player bending down with a glove.
    \item A bag of strawberries on a table with tomatoes.
    \item A crow standing on a plant in a body of water.
    \item A black-and-white cat sitting on a bed.
    \item A brown-and-white cow standing on a grassy hill.
    \item A brown leather couch in a living room.
    \item A body of water with an elephant in the background.
    \item A case containing a small doll with blue hair, shoes, and clothes.
\end{enumerate}

\noindent
\paragraph{Fig.~\ref{fig:4step-lpo} captions from left to right, top to bottom are:}
\begin{enumerate}
    \item A bowl of apples and bananas sitting on a woven cloth.
    \item A cake decorated with a surfer and palm trees.
    \item A close-up of a person eating a doughnut.
    \item A brown teddy bear sitting at a table next to a cup of coffee.
    \item A close-up of a metallic elephant statue.
    \item A baby in a gray jacket eating a piece of pizza crust.
    \item A couch and chair sitting in a room.
    \item A bench in the park on a rainy day.
    \item A brown-and-white cat sitting on a windowsill.
    \item A bedsit with a kitchenette featuring white cabinets.
    \item A bottle of wine placed next to a glass of wine.
    \item A blurred motorcycle against a red brick wall.
\end{enumerate}

\noindent
\paragraph{Fig.~\ref{fig:8step} captions from left to right, top to bottom are:}
\begin{enumerate}
    \item A bird on a plank looking at green water.
    \item A boy holding a pitcher’s mitt at a park.
    \item A clock near the front of a house.
    \item A bench sitting atop a lush green hillside.
    \item A black bear walking through a field of grass and straw.
    \item A black motorcycle parked on gray cobblestones.
    \item A child in bed wearing a striped sweater and colorful blanket.
    \item A brass-colored vase with a flower bouquet in it.
    \item A clean bedroom with a dog on the bed.
    \item A chef preparing sushi on a countertop.
    \item A close-up of a sandwich with French fries.
    \item A close-up image of a cat and a keyboard.
\end{enumerate}

\noindent
\paragraph{Fig.~\ref{fig:8step-lpo} captions from left to right, top to bottom are:}
\begin{enumerate}
    \item A bench along a sidewalk in winter, covered in snow.
    \item A man holding a little blond girl.
    \item A confection of cake, whipped cream, strawberries, and candles.
    \item A bedroom with a silky bedspread and pillows.
    \item A cat lying on a sofa next to some pillows.
    \item A brown teddy bear seated on a chair beside a wooden drawer.
    \item A baseball player standing next to home plate.
    \item A sleek motorcycle in a cityscape.
    \item A beautiful vase full of flowers, with pictures placed beside it.
    \item A beautiful bird standing on the bank of a river.
    \item A bagel topped with egg and other ingredients on a plate.
    \item A furry kitten lying on a laptop.
\end{enumerate}

\noindent
\paragraph{Fig.~\ref{fig:flashdmd-4step-sd3} captions from left to right, top to bottom are:}
\begin{enumerate}
    \item A mustard-yellow armchair with button tufting sits on a speckled white floor.  The chair has dark wooden legs.  A large potted plant is partially visible to the left.  A window with dark frames shows green trees and foliage.  A white rock sits on the floor near the chair. The floor is partially shaded by the window.
    \item A brown deer with large, curved antlers stands in a forest setting. The deer's coat is a uniform brown color.  Its antlers are dark brown and have multiple points. The deer's face is partially visible, showing its nose and eyes. The background is blurred with green foliage. The deer appears to be looking directly at the camera.
    \item Several cupcakes are arranged on a wire rack. One cupcake has dark blue frosting, gold sprinkles, and a jack-o'-lantern topper. Other cupcakes have orange frosting and are decorated with orange and purple sprinkles. The cupcakes are in purple and white patterned wrappers. The background is blurred.
    \item A doll with brown curly hair, blue eyes, and rosy cheeks wears a large black velvet hat with gold embroidery.  The doll's dress is black with gold trim and a white scarf tied around its neck.  The doll's face is white with painted features.  Other dolls are visible in the background. The doll appears to be wearing a black velvet dress with gold trim. The doll's hat has a wide brim and a decorative band.
    \item A vibrant orange rose is the focal point, its petals slightly unfurled. The rose is attached to a green stem with small thorns. The background is a blurred green foliage. The lighting is soft and natural. The rose appears to be in a natural setting.
    \item A young woman with fair skin, dark brown hair styled in an updo with a decorative hairpiece, wears a light-colored, sheer, embroidered traditional East Asian garment. She looks directly at the camera with a slight smile. The garment has floral embroidery in white and light blue. She wears small, dangling earrings.
    \item A green and pink parrot with a red beak eats a passion fruit. The parrot's head and neck are predominantly green, with a pink band around the neck. The beak is red and orange. The passion fruit is dark purple with a yellow interior and black seeds.
    \item A reddish-orange mushroom with a white speckled stem grows in a forest floor covered with fallen leaves and grass. A small, light brown leaf rests on the mushroom's cap. The background features blurred green foliage and trees. The lighting is soft and diffused.
    \item A white, fluffy dog lies in a field of green grass. The dog has its tongue out and is wearing a collar with a blue bone-shaped tag. The sun is low in the sky, creating a backlight.  The dog's fur appears slightly ruffled.  The grass is tall and slightly blurred.  Trees are visible in the background. The sky is a gradient of blue and light yellow.
    \item A bee is perched on a vibrant orange flower.  The flower has multiple petals and a yellow center.  Several other orange flowers and green leaves are visible in the background, some out of focus. The bee appears to be collecting nectar. The flowers are in a garden setting. The image is brightly lit, highlighting the orange hues of the flowers.
    \item A clear glass teapot containing a yellow liquid sits on a round wooden tray.  Several ripe red strawberries with green stems are placed on the tray alongside the teapot. The tray rests on a light brown, textured surface.  A blurred background of green foliage suggests an outdoor setting. The teapot has a copper-colored handle and lid. The light is bright and natural.
    \item A black mug with a gold design sits on a gray textured surface. The design depicts stylized mountains, trees, a crescent moon, and a campfire.  The words "Mountain Kind" are written below the design. Steam rises from the mug.  A lit candle, pine cones, cinnamon sticks, and a string of lights are also present. The background is blurred and dark.
\end{enumerate}

%% file: algorithm/stage1.tex
\begin{algorithm*}[htbp]
\caption{Flash-DMD Training Algorithm}
\label{alg:stage1}
\begin{algorithmic}[1]
\Require pretrained teacher model $\mu_{\text{real}}$, real dataset $\mathcal{D}_{\text{real}}$, generator is updated with the ratio $\texttt{TTUR}$, inference steps $K$, timestep set $S = \{\tau_1, \dots, \tau_k\}$ and its noisy set $S_\text{noisy}$, high noisy timesteps $T_\text{noisy}$, Pixel-level discriminator $D_{\omega}$, VAE decoder layers $\mathcal{V}$, latent reward model $R$, training stage flag $\texttt{FLAG}$.
\Ensure trained few-step generator $\mathcal{G}_{\theta}$

\State $\mathcal{G}_{\theta} \gets \text{copyWeights}(\mu_{\text{real}})$ \Comment{Initialize generator}
\State $\mu_{\phi} \gets \text{copyWeights}(\mu_{\text{real}})$ \Comment{Initialize score estimnator of generator}
\State $D_{\omega} \gets \text{initializeTrainableHeads}()$ \Comment{Initialize trainable heads of discriminator}

\For{iteration $= 1$ to max\_iters}
    \State $z \sim \mathcal{N}(0, I)$
    \State Sample $\tau_i$ from $S$ \Comment{Pick timestep for current iteration}
    \State Sample $x_{\text{real}} \sim \mathcal{D}_{\text{real}}$
    
    \State $x_{\tau_i} \gets \text{backwardSimulation}(z, \tau_k \to \tau_i)$ \Comment{Use backward simulation to get noisy image}
    \State $x_{\tau_1} \gets \text{backwardSimulation}(x_{\tau_i}, \tau_i \to \tau_1)$ \Comment{Use backward simulation to get clean image}

    \State $x \gets \mathcal{G}_{\theta}(x_{\tau_i}, \tau_i)$
    \State $p_\text{real}, p_\text{fake} \gets \mathcal{D}_{\omega}(\mathcal{V}(x_{\tau_1}))$ \Comment{Get real or fake probability}
    
    \If{iteration mod $\texttt{TTUR} == 0$}
        \State $t_{j} \gets T_\text{noisy}$
        \State $\mathcal{L}_{\text{DMD}} \gets \text{distributionMatchingLoss}(\mu_{\text{real}}, \mu_{\omega}, x, t_{j})$ \Comment{Use Eq.~\eqref{eq:timestep_aware_gen} for faster convergence in noisy timesteps}
        \State $\mathcal{L}_{\text{adv}} \gets \text{generatorAdversarialLoss}(p_\text{real})$ \Comment{Use Eq.~\eqref{eq:pixelgan_disc} for enhanced realism and details}
        
        \State $\mathcal{L}_{\mathcal{G}_{\theta}} \gets \mathcal{L}_{\text{DMD}} + \lambda \cdot \mathcal{L}_{\text{adv}}$ \Comment{Final loss function for generator}
        \State $\mathcal{G}_{\theta} \gets \text{update}(\mathcal{G}_{\theta}, \mathcal{L}_{\mathcal{G}_{\theta}})$ 
        \State $\mu_{\phi} \gets \text{EMA}(\theta, {\phi}, \lambda_{\text{ema1}})$
    \EndIf
    \State $x \gets x.\text{detach}()$ \Comment{Stop gradient}
    \State $t \sim \mathcal{U}(0, 1)$
    \State $x_t \gets \text{forwardDiffusion}(x, t)$ \Comment{Add noise}
    
    \State $\mathcal{L}_{\text{denoise}} \gets \text{diffusionLoss}(\mu_\phi(x_t, t), x)$
    \State $\mu_{\phi} \gets \text{update}(\mu_{\phi}, \mathcal{L}_{\text{denoise}})$ \Comment{Update fake score network $\mu_{\text{fake}}$}
    
    \State $\mathcal{L}_{D_{\omega}} \gets \text{discriminatorAdversarialLoss}(p_\text{real},p_\text{fake})$ \Comment{Discriminator's Hinge loss}
    \State $D_{\omega} \gets \text{update}(D_{\omega}, \mathcal{L}_{D_{\omega}})$ \Comment{Update discriminator $D_\omega$}

    \If{\texttt{FLAG} == \texttt{Stage2}} \Comment{Use reinforcement learning at the second stage}
    \State $s_\text{pool} \gets \text{iterativeSample}(S_\text{noisy}, K)$ \Comment{Sample K image latent for reward model evaluation}
    \State $s_\text{win}, s_\text{loss} \gets \text{filter}(\mathcal{R}(s_\text{pool}))$ \Comment{Construct win-loss pairs with $\mathcal{R}$}
    \State $\mathcal{L}_{rl} \gets \text{prefenceOptimization}(s_\text{win},s_\text{loss})$ \Comment{Use Eq.~\eqref{lpo} to boost performance}
    \State $\mathcal{G}_{\theta} \gets \text{update}(\mathcal{G}_{\theta}, \mathcal{L}_{rl})$ 
    \EndIf
\EndFor
\end{algorithmic}
\end{algorithm*}

%% file: tables/sdxl_8steps.tex
\definecolor{lightgray}{gray}{0.9}
\definecolor{lightblue}{RGB}{230, 240, 255}

\begin{table}[t]
\centering
\setlength{\tabcolsep}{2pt}
\caption{Comparison of \ours~on SDXL under stage 1 with other distillation methods on the COCO-10k dataset. \textbf{ImgRwd} denotes ImageReward score. Best performance is highlighted in \textbf{bold}, and the second best is \underline{underlined}.}
\vspace{-2mm}
\label{tab:sdxl8steps}
\resizebox{0.48\textwidth}{!}{
\begin{tabular}{lccccccc}
\toprule
\textbf{Method} & \textbf{\#NFE} & \textbf{ImgRwd} \updownarrows{$\uparrow$} & \textbf{CLIP} \updownarrows{$\uparrow$} & \textbf{Pick} \updownarrows{$\uparrow$} & \textbf{HPSv2} \updownarrows{$\uparrow$} & \textbf{MPS} \updownarrows{$\uparrow$} & \textbf{Cost} \updownarrows{$\downarrow$} \\
\midrule
\rowcolor{lightgray}
SDXL & 100 & 0.7143 & 0.3295 & 0.2265 & 0.2865 & 11.87 & - \\
\midrule
LCM-SDXL & 8 & 0.6122 & 0.3247 & 0.2261 & 0.2874 & 11.59 & - \\
SDXL-Lightning & 8 & 0.7187 & 0.3268 & 0.2291 & 0.2900 & 12.12 & - \\
Hyper-SD & 8 & 0.9119 & 0.3287 & 0.2310 & 0.2977 & 12.35 & - \\
\midrule
\multicolumn{8}{c}{\textbf{8-steps \ours~at Stage 1}} \\
\midrule
\rowcolor{lightblue}
TTUR2-3k & 8 & \underline{0.9159} & 0.3281 & \textbf{0.2319} & \underline{0.2981} & \underline{12.60} & 48*3k \\
TTUR2-6k & 8 & \textbf{0.9416} & 0.3284 & \underline{0.2318} & \textbf{0.2989} & \textbf{12.63} & 48*6k \\
\bottomrule
\end{tabular}
}
\end{table}

%% file: tables/sdxl_8steps_stage2.tex
\definecolor{lightblue}{RGB}{230, 240, 255}
\begin{table}[t]
\centering
\setlength{\tabcolsep}{3pt}
\caption{Comparison of \ours~under stage2 with other models with reinforcement learning on COCO-10k dataset.}
\vspace{-2mm}
\label{tab:sdxl8steps_stage2}
\resizebox{0.48\textwidth}{!}{
\begin{tabular}{lccccccc}
\toprule
\textbf{Method} & \textbf{\#NFE} & \textbf{ImgRwd} \updownarrows{$\uparrow$} & \textbf{CLIP} \updownarrows{$\uparrow$} & \textbf{Pick} \updownarrows{$\uparrow$} & \textbf{HPSv2} \updownarrows{$\uparrow$} & \textbf{MPS} \updownarrows{$\uparrow$} & \textbf{GPU Hours}\\
\midrule
Hyper-SDXL & 8 & 0.9119 & 0.3287 & 0.2310 & 0.2977 & 12.35 & 200 A100 \\
LPO-SDXL & 40 & \textbf{1.0417} & \textbf{0.3324} & \underline{0.2342} & {0.2965} & {12.58} & {92 A100} \\
\midrule
\multicolumn{8}{c}{\textbf{8-steps \ours~(TTUR2-3k) at Stage 2}} \\
\midrule
\rowcolor{lightblue}
\ours-3k & 8 & 1.0012 & \underline{0.3299} & 0.2338 & \underline{0.2986} & \underline{12.75} & \textbf{12 H20} \\
\ours-6k & 8 & \underline{1.0106} & {0.3290} & \textbf{0.2343} & \textbf{0.2998} & \textbf{12.84} & \underline{24 H20} \\
\bottomrule
\end{tabular}
}
\end{table}

%% file: visuals/visualization.tex
\begin{figure*}
    \centering 
    \includegraphics[width=0.9\textwidth, angle=0]{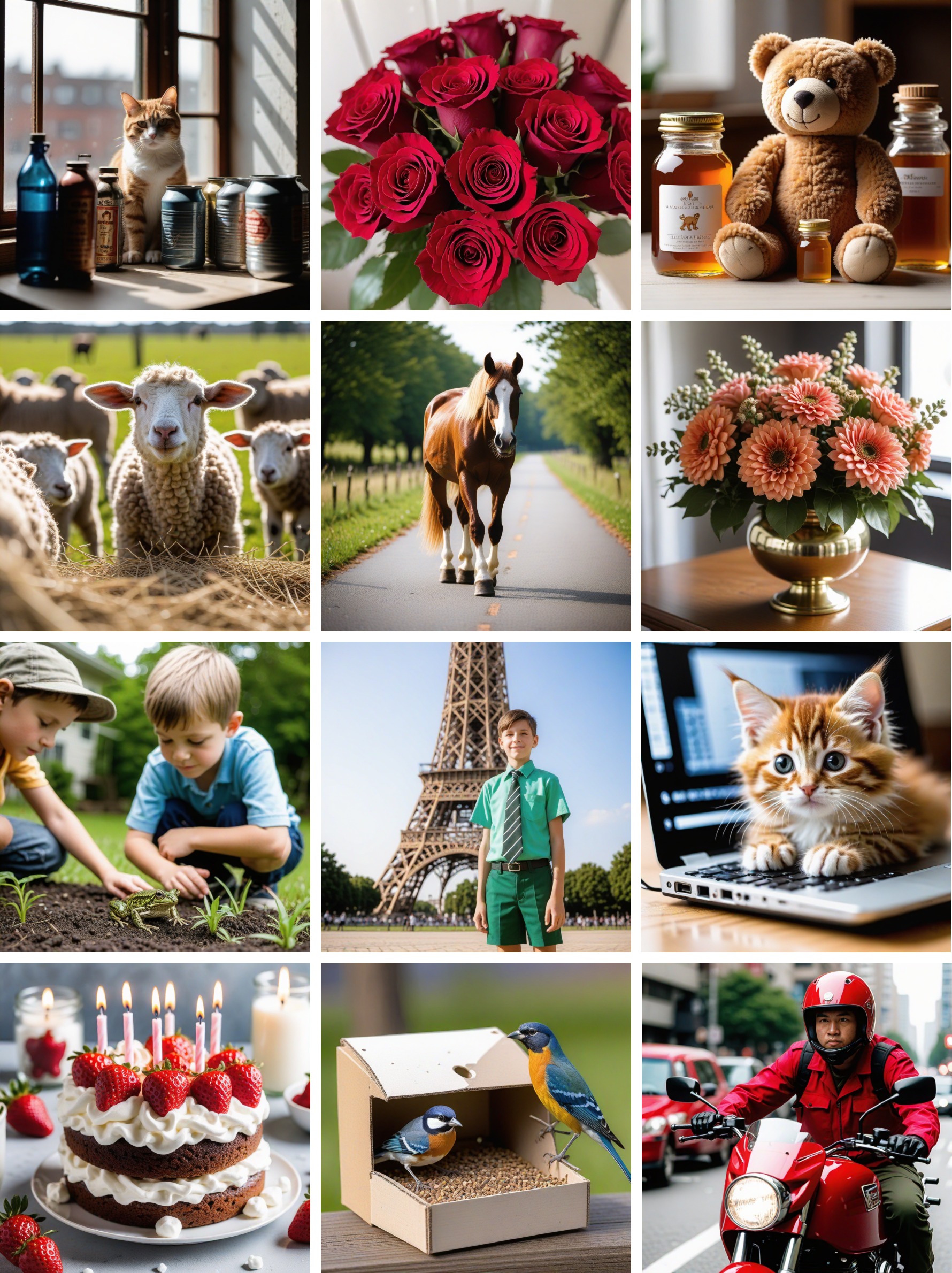}
    \caption{Qualitative results from Stage 1 of the 4-step Flash-DMD framework on SDXL. The model is trained with TTUR = 1 for 1,000 steps.}
    \label{fig:ttur1}
\end{figure*}

\begin{figure*}
    \centering 
    \includegraphics[width=0.85\textwidth, angle=0]{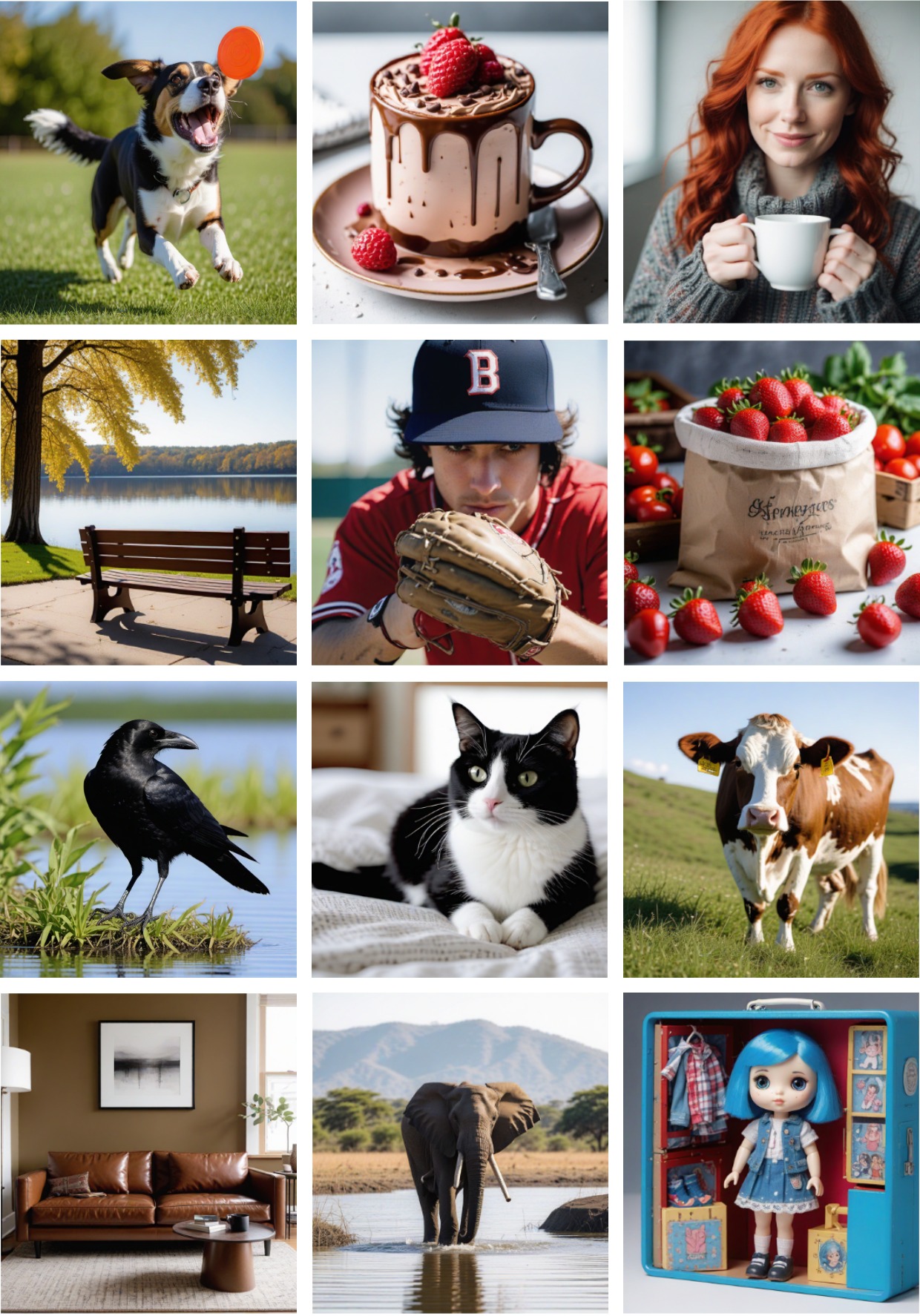}
    \caption{Qualitative results from Stage 1 of the 4-step Flash-DMD framework on SDXL. The model is trained with TTUR = 2 for 4,000 steps.}
    \label{fig:ttur2}
\end{figure*}

\begin{figure*}
    \centering 
    \includegraphics[width=0.9\textwidth, angle=0]{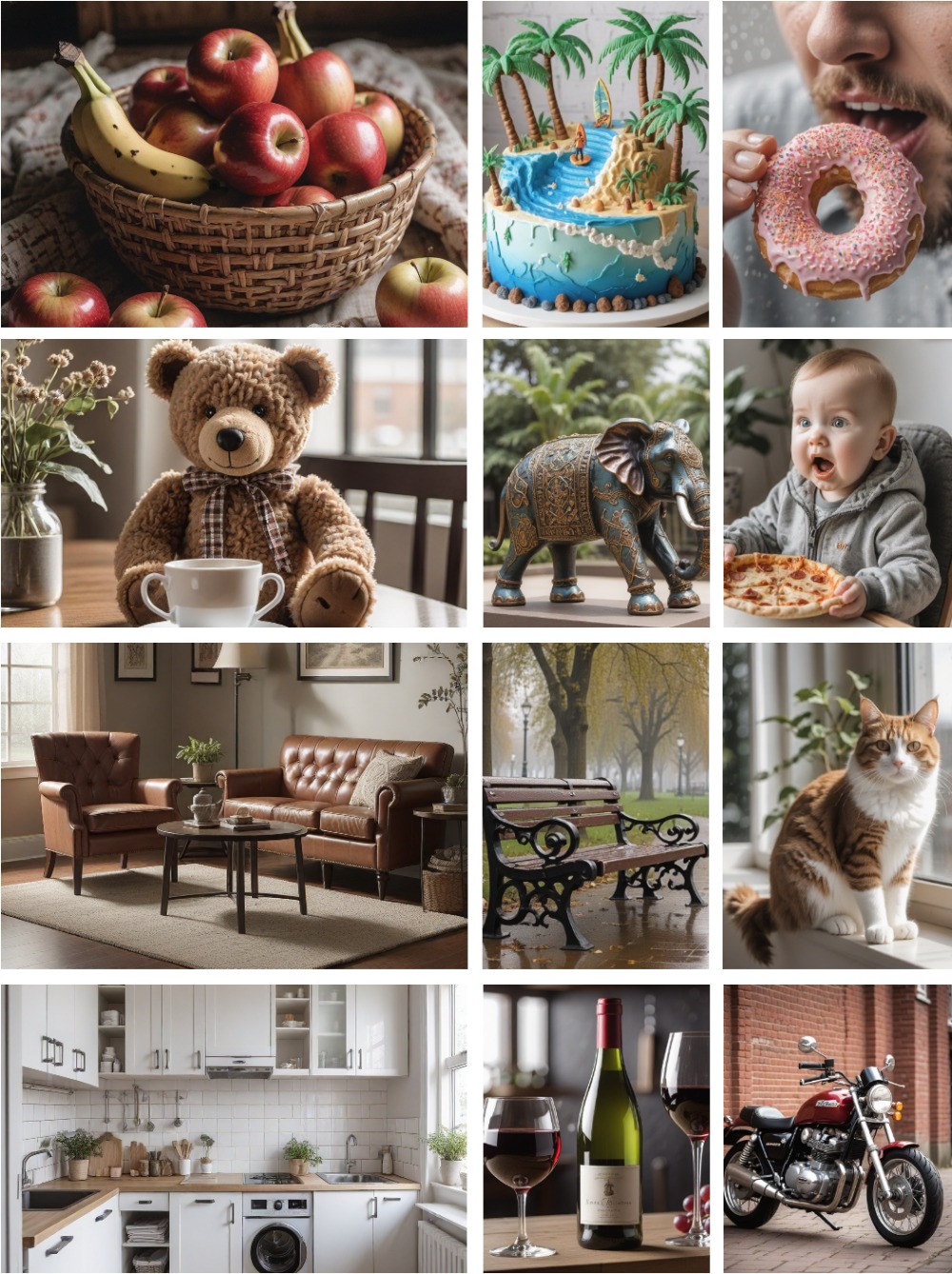}
    \caption{Qualitative results from Stage 2 of the 4-step Flash-DMD framework on SDXL. The model is initialized from the TTUR1-1k checkpoint and fine-tuned for 5,000 steps.}
    \label{fig:4step-lpo}
\end{figure*}

\begin{figure*}
    \centering 
    \includegraphics[width=0.9\textwidth, angle=0]{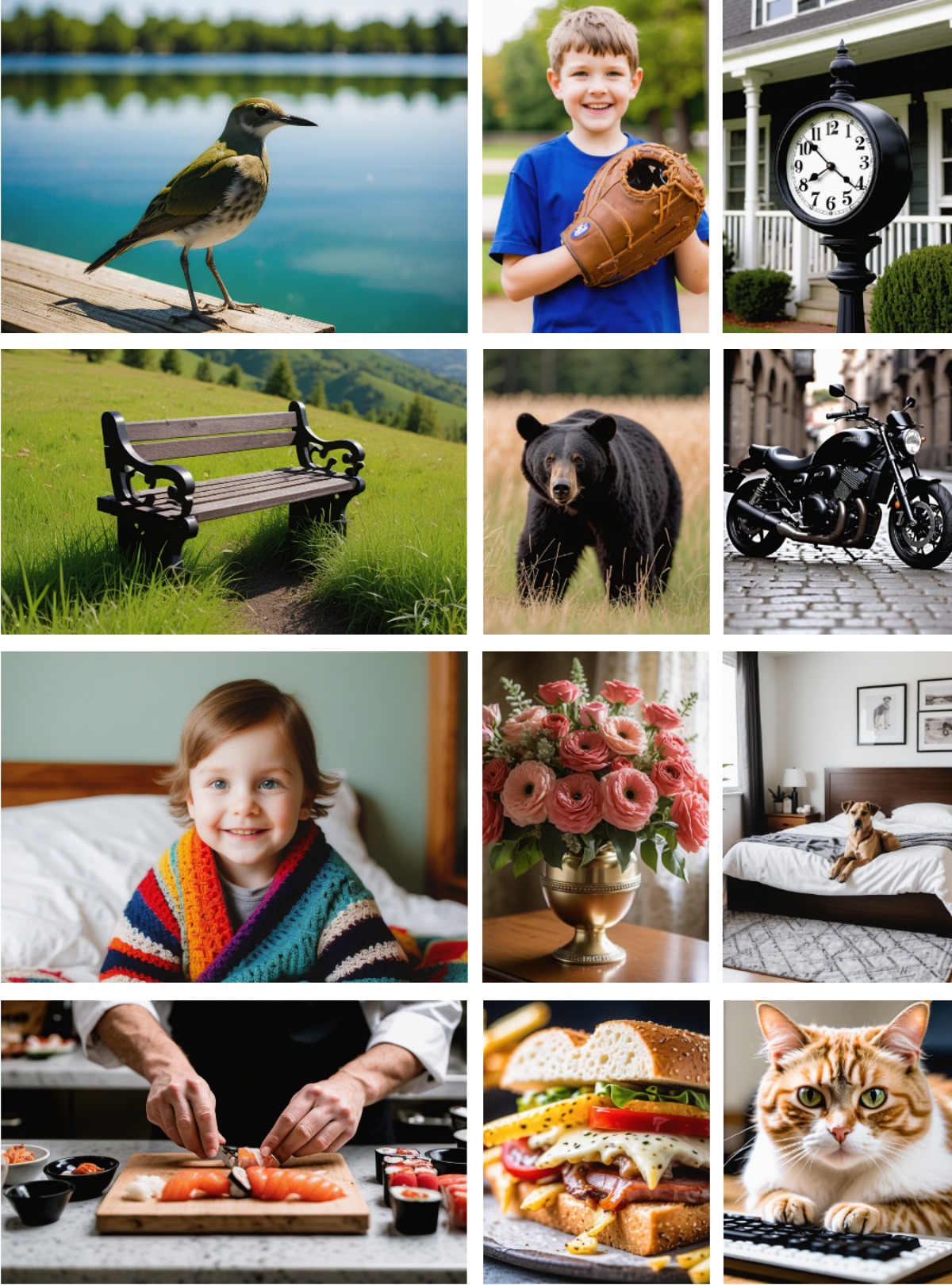}
    \caption{Qualitative results from Stage 1 of the 8-step Flash-DMD framework on SDXL. The model is trained with TTUR = 2 for 3,000 steps.}
    \label{fig:8step}
\end{figure*}

\begin{figure*}
    \centering 
    \includegraphics[width=0.85\textwidth, angle=0]{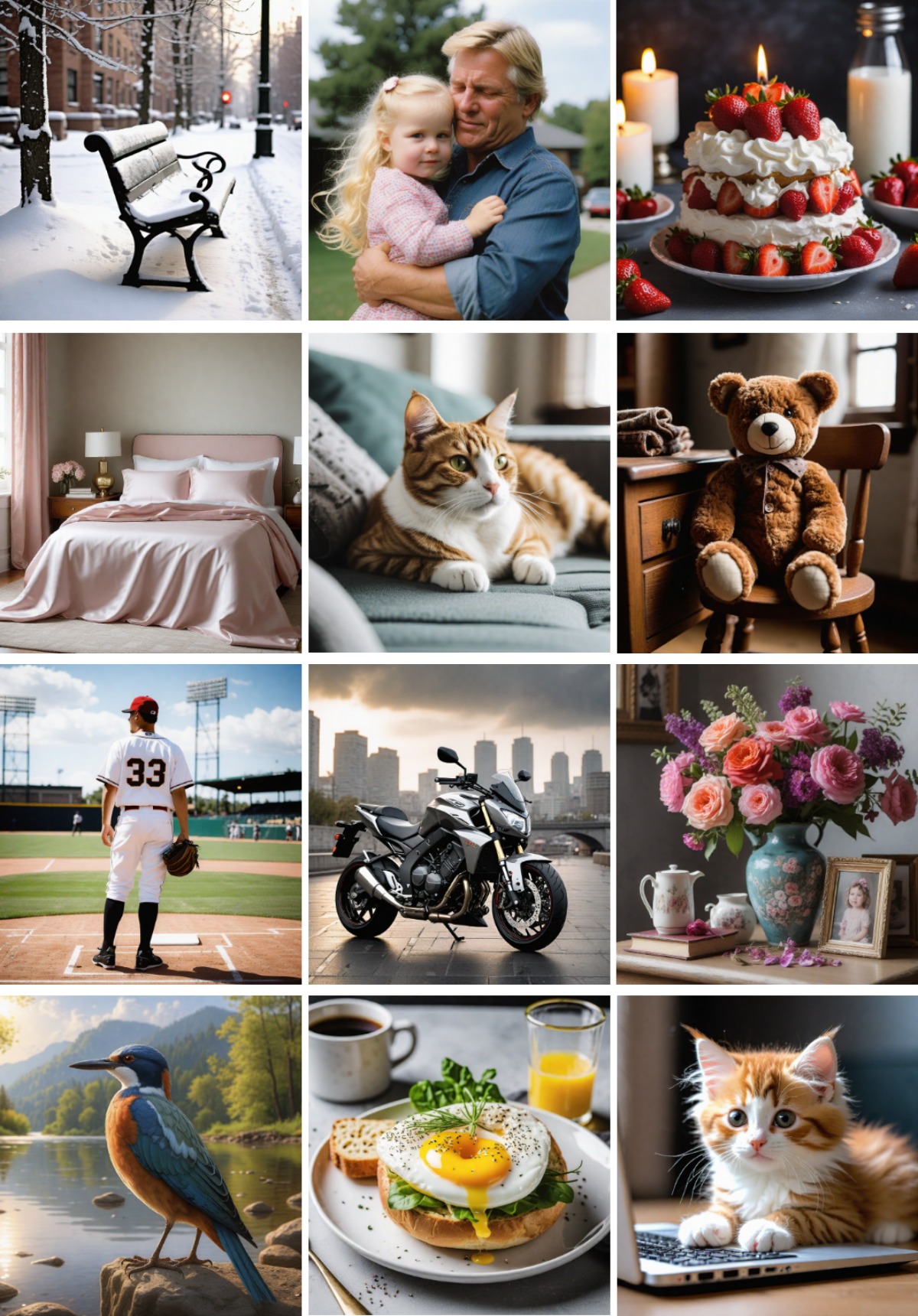}
    \caption{Qualitative results from Stage 2 of the 8-step Flash-DMD framework on SDXL. The model is initialized from the 8-step TTUR2-3k checkpoint and fine-tuned for 3,000 steps.}
    \label{fig:8step-lpo}
\end{figure*}

\begin{figure*}
    \centering
    \includegraphics[width=0.85\linewidth]{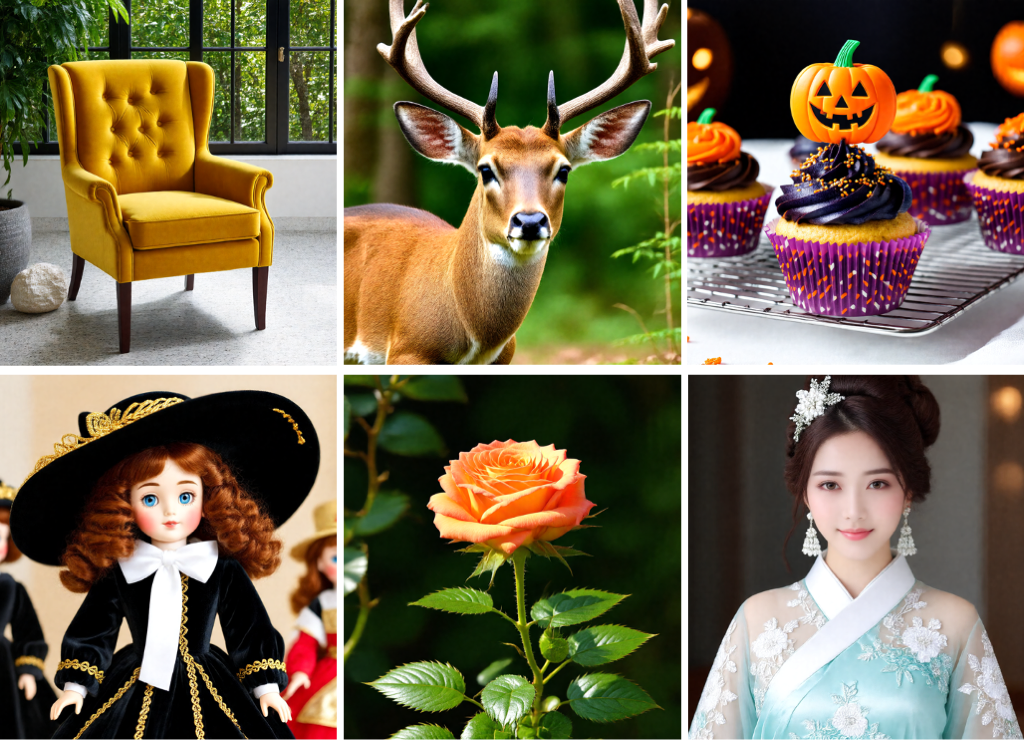}
    \includegraphics[width=0.85\linewidth]{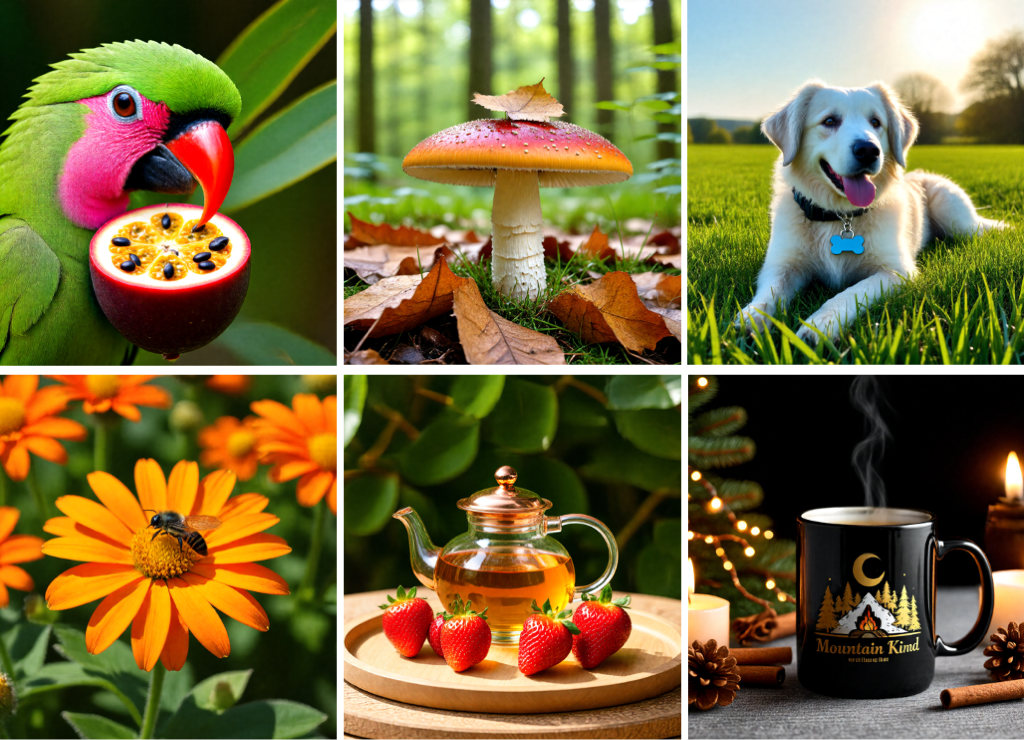}
    \caption{Qualitative results from stage 1 of the 4-step Flash-DMD framework on SD3-Medium. The model is trained with TTUR=2 for 7.000 steps.}
    \label{fig:flashdmd-4step-sd3}
\end{figure*}

%% file: main.bib
@inproceedings{DMD2,
    title={Improved Distribution Matching Distillation for Fast Image Synthesis},
    author={Yin, Tianwei and Gharbi, Micha{\"e}l and Park, Taesung and Zhang, Richard and Shechtman, Eli and Durand, Fredo and Freeman, William T},
    booktitle={NeurIPS},
    year={2024}
}

@article{zhang2025survey,
  title={A survey of reinforcement learning for large reasoning models},
  author={Zhang, Kaiyan and Zuo, Yuxin and He, Bingxiang and Sun, Youbang and Liu, Runze and Jiang, Che and Fan, Yuchen and Tian, Kai and Jia, Guoli and Li, Pengfei and others},
  journal={arXiv preprint arXiv:2509.08827},
  year={2025}
}

@inproceedings{DMD,
    title={One-step Diffusion with Distribution Matching Distillation},
    author={Yin, Tianwei and Gharbi, Micha{\"e}l and Zhang, Richard and Shechtman, Eli and Durand, Fr{\'e}do and Freeman, William T and Park, Taesung},
    booktitle={CVPR},
    year={2024}
}

@article{ADM,
  author       = {Yanzuo Lu and
                  Yuxi Ren and
                  Xin Xia and
                  Shanchuan Lin and
                  Xing Wang and
                  Xuefeng Xiao and
                  Andy J. Ma and
                  Xiaohua Xie and
                  Jian{-}Huang Lai},
  title        = {Adversarial Distribution Matching for Diffusion Distillation Towards
                  Efficient Image and Video Synthesis},
  journal      = {CoRR},
  volume       = {abs/2507.18569},
  year         = {2025},
  url          = {https://doi.org/10.48550/arXiv.2507.18569},
  doi          = {10.48550/ARXIV.2507.18569},
  eprinttype    = {arXiv},
  eprint       = {2507.18569},
  bibsource    = {dblp computer science bibliography, https://dblp.org}
}

@article{senseflow,
  author       = {Xingtong Ge and
                  Xin Zhang and
                  Tongda Xu and
                  Yi Zhang and
                  Xinjie Zhang and
                  Yan Wang and
                  Jun Zhang},
  title        = {SenseFlow: Scaling Distribution Matching for Flow-based Text-to-Image
                  Distillation},
  journal      = {CoRR},
  volume       = {abs/2506.00523},
  year         = {2025},
  url          = {https://doi.org/10.48550/arXiv.2506.00523},
  doi          = {10.48550/ARXIV.2506.00523},
  eprinttype    = {arXiv},
  eprint       = {2506.00523},
  bibsource    = {dblp computer science bibliography, https://dblp.org}
}

@article{diffusion,
  title={Denoising diffusion probabilistic models},
  author={Ho, Jonathan and Jain, Ajay and Abbeel, Pieter},
  journal={Advances in neural information processing systems},
  volume={33},
  pages={6840--6851},
  year={2020}
}

@inproceedings{latentdiffusion,
  title={High-resolution image synthesis with latent diffusion models},
  author={Rombach, Robin and Blattmann, Andreas and Lorenz, Dominik and Esser, Patrick and Ommer, Bj{\"o}rn},
  booktitle={Proceedings of the IEEE/CVF conference on computer vision and pattern recognition},
  pages={10684--10695},
  year={2022}
}

@article{sdxl,
  title={Sdxl: Improving latent diffusion models for high-resolution image synthesis},
  author={Podell, Dustin and English, Zion and Lacey, Kyle and Blattmann, Andreas and Dockhorn, Tim and M{\"u}ller, Jonas and Penna, Joe and Rombach, Robin},
  journal={arXiv preprint arXiv:2307.01952},
  year={2023}
}

@inproceedings{flowdiffusion,
  title={Scaling rectified flow transformers for high-resolution image synthesis},
  author={Esser, Patrick and Kulal, Sumith and Blattmann, Andreas and Entezari, Rahim and M{\"u}ller, Jonas and Saini, Harry and Levi, Yam and Lorenz, Dominik and Sauer, Axel and Boesel, Frederic and others},
  booktitle={Forty-first international conference on machine learning},
  year={2024}
}

@misc{flux2024,
    author={Black Forest Labs},
    title={FLUX},
    year={2024},
    howpublished={\url{https://github.com/black-forest-labs/flux}},
}

@misc{visionreward,
      title={VisionReward: Fine-Grained Multi-Dimensional Human Preference Learning for Image and Video Generation}, 
      author={Jiazheng Xu and Yu Huang and Jiale Cheng and Yuanming Yang and Jiajun Xu and Yuan Wang and Wenbo Duan and Shen Yang and Qunlin Jin and Shurun Li and Jiayan Teng and Zhuoyi Yang and Wendi Zheng and Xiao Liu and Ming Ding and Xiaohan Zhang and Xiaotao Gu and Shiyu Huang and Minlie Huang and Jie Tang and Yuxiao Dong},
      year={2024},
      eprint={2412.21059},
      archivePrefix={arXiv},
      primaryClass={cs.CV},
      url={https://arxiv.org/abs/2412.21059}, 
}

@inproceedings{clip,
  title={Learning transferable visual models from natural language supervision},
  author={Radford, Alec and Kim, Jong Wook and Hallacy, Chris and Ramesh, Aditya and Goh, Gabriel and Agarwal, Sandhini and Sastry, Girish and Askell, Amanda and Mishkin, Pamela and Clark, Jack and others},
  booktitle={International conference on machine learning},
  pages={8748--8763},
  year={2021},
  organization={PmLR}
}

@article{bai2025qwen2,
  title={Qwen2. 5-vl technical report},
  author={Bai, Shuai and Chen, Keqin and Liu, Xuejing and Wang, Jialin and Ge, Wenbin and Song, Sibo and Dang, Kai and Wang, Peng and Wang, Shijie and Tang, Jun and others},
  journal={arXiv preprint arXiv:2502.13923},
  year={2025}
}

@inproceedings{li2022blip,
  title={Blip: Bootstrapping language-image pre-training for unified vision-language understanding and generation},
  author={Li, Junnan and Li, Dongxu and Xiong, Caiming and Hoi, Steven},
  booktitle={International conference on machine learning},
  pages={12888--12900},
  year={2022},
  organization={PMLR}
}

@misc{LCM,
      title={Latent Consistency Models: Synthesizing High-Resolution Images with Few-Step Inference}, 
      author={Simian Luo and Yiqin Tan and Longbo Huang and Jian Li and Hang Zhao},
      year={2023},
      eprint={2310.04378},
      archivePrefix={arXiv},
      primaryClass={cs.CV}
}

@article{lcmlora,
  title={Lcm-lora: A universal stable-diffusion acceleration module},
  author={Luo, Simian and Tan, Yiqin and Patil, Suraj and Gu, Daniel and von Platen, Patrick and Passos, Apolin{\'a}rio and Huang, Longbo and Li, Jian and Zhao, Hang},
  journal={arXiv preprint arXiv:2311.05556},
  year={2023}
}

@inproceedings{ADD,
  title={Adversarial diffusion distillation},
  author={Sauer, Axel and Lorenz, Dominik and Blattmann, Andreas and Rombach, Robin},
  booktitle={European Conference on Computer Vision},
  pages={87--103},
  year={2024},
  organization={Springer}
}

@inproceedings{LADD,
  title={Fast high-resolution image synthesis with latent adversarial diffusion distillation},
  author={Sauer, Axel and Boesel, Frederic and Dockhorn, Tim and Blattmann, Andreas and Esser, Patrick and Rombach, Robin},
  booktitle={SIGGRAPH Asia 2024 Conference Papers},
  pages={1--11},
  year={2024}
}

@article{sdxllighting,
  title={Sdxl-lightning: Progressive adversarial diffusion distillation},
  author={Lin, Shanchuan and Wang, Anran and Yang, Xiao},
  journal={arXiv preprint arXiv:2402.13929},
  year={2024}
}

@article{hypersd,
  title={Hyper-sd: Trajectory segmented consistency model for efficient image synthesis},
  author={Ren, Yuxi and Xia, Xin and Lu, Yanzuo and Zhang, Jiacheng and Wu, Jie and Xie, Pan and Wang, Xing and Xiao, Xuefeng},
  journal={Advances in Neural Information Processing Systems},
  volume={37},
  pages={117340--117362},
  year={2024}
}

@article{PCM,
  title={Phased consistency models},
  author={Wang, Fu-Yun and Huang, Zhaoyang and Bergman, Alexander and Shen, Dazhong and Gao, Peng and Lingelbach, Michael and Sun, Keqiang and Bian, Weikang and Song, Guanglu and Liu, Yu and others},
  journal={Advances in neural information processing systems},
  volume={37},
  pages={83951--84009},
  year={2024}
}

@inproceedings{flashsdxl,
  title={Flash diffusion: Accelerating any conditional diffusion model for few steps image generation},
  author={Chadebec, Clement and Tasar, Onur and Benaroche, Eyal and Aubin, Benjamin},
  booktitle={Proceedings of the AAAI Conference on Artificial Intelligence},
  volume={39},
  number={15},
  pages={15686--15695},
  year={2025}
}

@article{flowgrpo,
  author       = {Jie Liu and
                  Gongye Liu and
                  Jiajun Liang and
                  Yangguang Li and
                  Jiaheng Liu and
                  Xintao Wang and
                  Pengfei Wan and
                  Di Zhang and
                  Wanli Ouyang},
  title        = {Flow-GRPO: Training Flow Matching Models via Online {RL}},
  journal      = {CoRR},
  volume       = {abs/2505.05470},
  year         = {2025},
  url          = {https://doi.org/10.48550/arXiv.2505.05470},
  doi          = {10.48550/ARXIV.2505.05470},
  eprinttype    = {arXiv},
  eprint       = {2505.05470},
  bibsource    = {dblp computer science bibliography, https://dblp.org}
}

@article{dancegrpo,
  author       = {Zeyue Xue and
                  Jie Wu and
                  Yu Gao and
                  Fangyuan Kong and
                  Lingting Zhu and
                  Mengzhao Chen and
                  Zhiheng Liu and
                  Wei Liu and
                  Qiushan Guo and
                  Weilin Huang and
                  Ping Luo},
  title        = {DanceGRPO: Unleashing {GRPO} on Visual Generation},
  journal      = {CoRR},
  volume       = {abs/2505.07818},
  year         = {2025},
  url          = {https://doi.org/10.48550/arXiv.2505.07818},
  doi          = {10.48550/ARXIV.2505.07818},
  eprinttype    = {arXiv},
  eprint       = {2505.07818},
  bibsource    = {dblp computer science bibliography, https://dblp.org}
}

@article{mixgrpo,
  title={Mixgrpo: Unlocking flow-based grpo efficiency with mixed ode-sde},
  author={Li, Junzhe and Cui, Yutao and Huang, Tao and Ma, Yinping and Fan, Chun and Yang, Miles and Zhong, Zhao},
  journal={arXiv preprint arXiv:2507.21802},
  year={2025}
}

@article{prefgrpo,
  title={Pref-GRPO: Pairwise Preference Reward-based GRPO for Stable Text-to-Image Reinforcement Learning},
  author={Wang, Yibin and Li, Zhimin and Zang, Yuhang and Zhou, Yujie and Bu, Jiazi and Wang, Chunyu and Lu, Qinglin and Jin, Cheng and Wang, Jiaqi},
  journal={arXiv preprint arXiv:2508.20751},
  year={2025}
}

@article{tempflowgrpo,
  title={Tempflow-grpo: When timing matters for grpo in flow models},
  author={He, Xiaoxuan and Fu, Siming and Zhao, Yuke and Li, Wanli and Yang, Jian and Yin, Dacheng and Rao, Fengyun and Zhang, Bo},
  journal={arXiv preprint arXiv:2508.04324},
  year={2025}
}

@inproceedings{diffusiondpo,
  author       = {Bram Wallace and
                  Meihua Dang and
                  Rafael Rafailov and
                  Linqi Zhou and
                  Aaron Lou and
                  Senthil Purushwalkam and
                  Stefano Ermon and
                  Caiming Xiong and
                  Shafiq Joty and
                  Nikhil Naik},
  title        = {Diffusion Model Alignment Using Direct Preference Optimization},
  booktitle    = {{IEEE/CVF} Conference on Computer Vision and Pattern Recognition,
                  {CVPR} 2024, Seattle, WA, USA, June 16-22, 2024},
  pages        = {8228--8238},
  publisher    = {{IEEE}},
  year         = {2024},
  url          = {https://doi.org/10.1109/CVPR52733.2024.00786},
  doi          = {10.1109/CVPR52733.2024.00786},
  bibsource    = {dblp computer science bibliography, https://dblp.org}
}

@inproceedings{spo,
  author       = {Zhanhao Liang and
                  Yuhui Yuan and
                  Shuyang Gu and
                  Bohan Chen and
                  Tiankai Hang and
                  Mingxi Cheng and
                  Ji Li and
                  Liang Zheng},
  title        = {Aesthetic Post-Training Diffusion Models from Generic Preferences
                  with Step-by-step Preference Optimization},
  booktitle    = {{IEEE/CVF} Conference on Computer Vision and Pattern Recognition,
                  {CVPR} 2025, Nashville, TN, USA, June 11-15, 2025},
  pages        = {13199--13208},
  publisher    = {Computer Vision Foundation / {IEEE}},
  year         = {2025},
  url          = {https://openaccess.thecvf.com/content/CVPR2025/html/Liang\_Aesthetic\_Post-Training\_Diffusion\_Models\_from\_Generic\_Preferences\_with\_Step-by-step\_Preference\_CVPR\_2025\_paper.html},
  doi          = {10.1109/CVPR52734.2025.01232},
  bibsource    = {dblp computer science bibliography, https://dblp.org}
}

@article{lpo,
  author       = {Tao Zhang and
                  Cheng Da and
                  Kun Ding and
                  Kun Jin and
                  Yan Li and
                  Tingting Gao and
                  Di Zhang and
                  Shiming Xiang and
                  Chunhong Pan},
  title        = {Diffusion Model as a Noise-Aware Latent Reward Model for Step-Level
                  Preference Optimization},
  journal      = {CoRR},
  volume       = {abs/2502.01051},
  year         = {2025},
  url          = {https://doi.org/10.48550/arXiv.2502.01051},
  doi          = {10.48550/ARXIV.2502.01051},
  eprinttype    = {arXiv},
  eprint       = {2502.01051},
  bibsource    = {dblp computer science bibliography, https://dblp.org}
}

@article{inversiondpo,
  author       = {Zejian Li and
                  Yize Li and
                  Chenye Meng and
                  Zhongni Liu and
                  Yang Ling and
                  Shengyuan Zhang and
                  Guang Yang and
                  Changyuan Yang and
                  Zhiyuan Yang and
                  Lingyun Sun},
  title        = {Inversion-DPO: Precise and Efficient Post-Training for Diffusion Models},
  journal      = {CoRR},
  volume       = {abs/2507.11554},
  year         = {2025},
  url          = {https://doi.org/10.48550/arXiv.2507.11554},
  doi          = {10.48550/ARXIV.2507.11554},
  eprinttype    = {arXiv},
  eprint       = {2507.11554},
  bibsource    = {dblp computer science bibliography, https://dblp.org}
}

@inproceedings{pso,
  author       = {Zichen Miao and
                  Zhengyuan Yang and
                  Kevin Lin and
                  Ze Wang and
                  Zicheng Liu and
                  Lijuan Wang and
                  Qiang Qiu},
  title        = {Tuning Timestep-Distilled Diffusion Model Using Pairwise Sample Optimization},
  booktitle    = {The Thirteenth International Conference on Learning Representations,
                  {ICLR} 2025, Singapore, April 24-28, 2025},
  publisher    = {OpenReview.net},
  year         = {2025},
  url          = {https://openreview.net/forum?id=fXnE4gB64o},
  bibsource    = {dblp computer science bibliography, https://dblp.org}
}

@inproceedings{capo,
  author       = {Kyungmin Lee and
                  Xiahong Li and
                  Qifei Wang and
                  Junfeng He and
                  Junjie Ke and
                  Ming{-}Hsuan Yang and
                  Irfan Essa and
                  Jinwoo Shin and
                  Feng Yang and
                  Yinxiao Li},
  title        = {Calibrated Multi-Preference Optimization for Aligning Diffusion Models},
  booktitle    = {{IEEE/CVF} Conference on Computer Vision and Pattern Recognition,
                  {CVPR} 2025, Nashville, TN, USA, June 11-15, 2025},
  pages        = {18465--18475},
  publisher    = {Computer Vision Foundation / {IEEE}},
  year         = {2025},
  url          = {https://openaccess.thecvf.com/content/CVPR2025/html/Lee\_Calibrated\_Multi-Preference\_Optimization\_for\_Aligning\_Diffusion\_Models\_CVPR\_2025\_paper.html},
  doi          = {10.1109/CVPR52734.2025.01721},
  bibsource    = {dblp computer science bibliography, https://dblp.org}
}

@article{scoredistill,
  title={Text-to-3d with classifier score distillation},
  author={Yu, Xin and Guo, Yuan-Chen and Li, Yangguang and Liang, Ding and Zhang, Song-Hai and Qi, Xiaojuan},
  journal={arXiv preprint arXiv:2310.19415},
  year={2023}
}

@inproceedings{dit,
  title={Scalable diffusion models with transformers},
  author={Peebles, William and Xie, Saining},
  booktitle={Proceedings of the IEEE/CVF international conference on computer vision},
  pages={4195--4205},
  year={2023}
}

@article{skipdit,
  title={Towards Stabilized and Efficient Diffusion Transformers through Long-Skip-Connections with Spectral Constraints},
  author={Chen, Guanjie and Zhao, Xinyu and Zhou, Yucheng and Qu, Xiaoye and Chen, Tianlong and Cheng, Yu},
  journal={arXiv preprint arXiv:2411.17616},
  year={2024}
}

@article{song2019generative,
  title={Generative modeling by estimating gradients of the data distribution},
  author={Song, Yang and Ermon, Stefano},
  journal={Advances in neural information processing systems},
  volume={32},
  year={2019}
}

@inproceedings{lin2014microsoft,
  title={Microsoft coco: Common objects in context},
  author={Lin, Tsung-Yi and Maire, Michael and Belongie, Serge and Hays, James and Perona, Pietro and Ramanan, Deva and Doll{\'a}r, Piotr and Zitnick, C Lawrence},
  booktitle={European conference on computer vision},
  pages={740--755},
  year={2014},
  organization={Springer}
}

@article{khan2022transformers,
  title={Transformers in vision: A survey},
  author={Khan, Salman and Naseer, Muzammal and Hayat, Munawar and Zamir, Syed Waqas and Khan, Fahad Shahbaz and Shah, Mubarak},
  journal={ACM computing surveys (CSUR)},
  volume={54},
  number={10s},
  pages={1--41},
  year={2022},
  publisher={ACM New York, NY}
}

@inproceedings{unet,
  title={U-net: Convolutional networks for biomedical image segmentation},
  author={Ronneberger, Olaf and Fischer, Philipp and Brox, Thomas},
  booktitle={International Conference on Medical image computing and computer-assisted intervention},
  pages={234--241},
  year={2015},
  organization={Springer}
}

@inproceedings{radford2021learning,
  title={Learning transferable visual models from natural language supervision},
  author={Radford, Alec and Kim, Jong Wook and Hallacy, Chris and Ramesh, Aditya and Goh, Gabriel and Agarwal, Sandhini and Sastry, Girish and Askell, Amanda and Mishkin, Pamela and Clark, Jack and others},
  booktitle={International conference on machine learning},
  pages={8748--8763},
  year={2021},
  organization={PmLR}
}

@inproceedings{MPS,
  title={Learning multi-dimensional human preference for text-to-image generation},
  author={Zhang, Sixian and Wang, Bohan and Wu, Junqiang and Li, Yan and Gao, Tingting and Zhang, Di and Wang, Zhongyuan},
  booktitle={Proceedings of the IEEE/CVF Conference on Computer Vision and Pattern Recognition},
  pages={8018--8027},
  year={2024}
}

@article{hpsv2,
  title={Human preference score v2: A solid benchmark for evaluating human preferences of text-to-image synthesis},
  author={Wu, Xiaoshi and Hao, Yiming and Sun, Keqiang and Chen, Yixiong and Zhu, Feng and Zhao, Rui and Li, Hongsheng},
  journal={arXiv preprint arXiv:2306.09341},
  year={2023}
}

@article{xu2023imagereward,
  title={Imagereward: Learning and evaluating human preferences for text-to-image generation},
  author={Xu, Jiazheng and Liu, Xiao and Wu, Yuchen and Tong, Yuxuan and Li, Qinkai and Ding, Ming and Tang, Jie and Dong, Yuxiao},
  journal={Advances in Neural Information Processing Systems},
  volume={36},
  pages={15903--15935},
  year={2023}
}

@article{pickscore,
  title={Pick-a-pic: An open dataset of user preferences for text-to-image generation},
  author={Kirstain, Yuval and Polyak, Adam and Singer, Uriel and Matiana, Shahbuland and Penna, Joe and Levy, Omer},
  journal={Advances in neural information processing systems},
  volume={36},
  pages={36652--36663},
  year={2023}
}

@article{yin2024improved,
  title={Improved distribution matching distillation for fast image synthesis},
  author={Yin, Tianwei and Gharbi, Micha{\"e}l and Park, Taesung and Zhang, Richard and Shechtman, Eli and Durand, Fredo and Freeman, Bill},
  journal={Advances in neural information processing systems},
  volume={37},
  pages={47455--47487},
  year={2024}
}

@article{schuhmann2022laion,
  title={Laion-5b: An open large-scale dataset for training next generation image-text models},
  author={Schuhmann, Christoph and Beaumont, Romain and Vencu, Richard and Gordon, Cade and Wightman, Ross and Cherti, Mehdi and Coombes, Theo and Katta, Aarush and Mullis, Clayton and Wortsman, Mitchell and others},
  journal={Advances in neural information processing systems},
  volume={35},
  pages={25278--25294},
  year={2022}
}

@article{salimans2022progressive,
  title={Progressive distillation for fast sampling of diffusion models},
  author={Salimans, Tim and Ho, Jonathan},
  journal={arXiv preprint arXiv:2202.00512},
  year={2022}
}

@article{tcm,
  title={Trajectory consistency distillation},
  author={Zheng, Jianbin and Hu, Minghui and Fan, Zhongyi and Wang, Chaoyue and Ding, Changxing and Tao, Dacheng and Cham, Tat-Jen},
  journal={CoRR},
  year={2024}
}

@inproceedings{progressive,
  author       = {Chenlin Meng and
                  Robin Rombach and
                  Ruiqi Gao and
                  Diederik P. Kingma and
                  Stefano Ermon and
                  Jonathan Ho and
                  Tim Salimans},
  title        = {On Distillation of Guided Diffusion Models},
  booktitle    = {{IEEE/CVF} Conference on Computer Vision and Pattern Recognition,
                  {CVPR} 2023, Vancouver, BC, Canada, June 17-24, 2023},
  pages        = {14297--14306},
  publisher    = {{IEEE}},
  year         = {2023},
  url          = {https://doi.org/10.1109/CVPR52729.2023.01374},
  doi          = {10.1109/CVPR52729.2023.01374},
  bibsource    = {dblp computer science bibliography, https://dblp.org}
}

@article{franceschi2023unifying,
  title={Unifying gans and score-based diffusion as generative particle models},
  author={Franceschi, Jean-Yves and Gartrell, Mike and Dos Santos, Ludovic and Issenhuth, Thibaut and de B{\'e}zenac, Emmanuel and Chen, Micka{\"e}l and Rakotomamonjy, Alain},
  journal={Advances in Neural Information Processing Systems},
  volume={36},
  pages={59729--59760},
  year={2023}
}

@inproceedings{nguyen2024swiftbrush,
  title={Swiftbrush: One-step text-to-image diffusion model with variational score distillation},
  author={Nguyen, Thuan Hoang and Tran, Anh},
  booktitle={Proceedings of the IEEE/CVF Conference on Computer Vision and Pattern Recognition},
  pages={7807--7816},
  year={2024}
}

@article{cheng2025pose,
  title={POSE: Phased One-Step Adversarial Equilibrium for Video Diffusion Models},
  author={Cheng, Jiaxiang and Ma, Bing and Ren, Xuhua and Jin, Hongyi and Yu, Kai and Zhang, Peng and Li, Wenyue and Zhou, Yuan and Zheng, Tianxiang and Lu, Qinglin},
  journal={arXiv preprint arXiv:2508.21019},
  year={2025}
}

@inproceedings{sam,
  title={Segment anything},
  author={Kirillov, Alexander and Mintun, Eric and Ravi, Nikhila and Mao, Hanzi and Rolland, Chloe and Gustafson, Laura and Xiao, Tete and Whitehead, Spencer and Berg, Alexander C and Lo, Wan-Yen and others},
  booktitle={Proceedings of the IEEE/CVF international conference on computer vision},
  pages={4015--4026},
  year={2023}
}

@inproceedings{nitro,
  author       = {Dar{-}Yen Chen and
                  Hmrishav Bandyopadhyay and
                  Kai Zou and
                  Yi{-}Zhe Song},
  title        = {NitroFusion: High-Fidelity Single-Step Diffusion through Dynamic Adversarial
                  Training},
  booktitle    = {{IEEE/CVF} Conference on Computer Vision and Pattern Recognition,
                  {CVPR} 2025, Nashville, TN, USA, June 11-15, 2025},
  pages        = {7654--7663},
  publisher    = {Computer Vision Foundation / {IEEE}},
  year         = {2025},
  url          = {https://openaccess.thecvf.com/content/CVPR2025/html/Chen\_NitroFusion\_High-Fidelity\_Single-Step\_Diffusion\_through\_Dynamic\_Adversarial\_Training\_CVPR\_2025\_paper.html},
  doi          = {10.1109/CVPR52734.2025.00717},
  bibsource    = {dblp computer science bibliography, https://dblp.org}
}

@article{zhang2025diffusion,
  title={Diffusion model as a noise-aware latent reward model for step-level preference optimization},
  author={Zhang, Tao and Da, Cheng and Ding, Kun and Yang, Huan and Jin, Kun and Li, Yan and Gao, Tingting and Zhang, Di and Xiang, Shiming and Pan, Chunhong},
  journal={arXiv preprint arXiv:2502.01051},
  year={2025}
}

@inproceedings{sd3,
  title={Scaling rectified flow transformers for high-resolution image synthesis},
  author={Esser, Patrick and Kulal, Sumith and Blattmann, Andreas and Entezari, Rahim and M{\"u}ller, Jonas and Saini, Harry and Levi, Yam and Lorenz, Dominik and Sauer, Axel and Boesel, Frederic and others},
  booktitle={Forty-first international conference on machine learning}
}

@inproceedings{chadebec2025flash,
  title={Flash diffusion: Accelerating any conditional diffusion model for few steps image generation},
  author={Chadebec, Clement and Tasar, Onur and Benaroche, Eyal and Aubin, Benjamin},
  booktitle={Proceedings of the AAAI Conference on Artificial Intelligence},
  volume={39},
  number={15},
  pages={15686--15695},
  year={2025}
}

@article{liao2025step,
  title={Step-level Reward for Free in RL-based T2I Diffusion Model Fine-tuning},
  author={Liao, Xinyao and Wei, Wei and Qu, Xiaoye and Cheng, Yu},
  journal={arXiv preprint arXiv:2505.19196},
  year={2025}
}
